\pdfoutput=1

\documentclass[11pt]{article}
\usepackage{booktabs}
\usepackage{multirow}
\usepackage{tabularx}

\usepackage{EMNLP2023}

\usepackage{times}
\usepackage{latexsym}
\usepackage{longtable}

\usepackage{hyperref}
\usepackage{nameref}

\usepackage[T1]{fontenc}

\usepackage[utf8]{inputenc}

\usepackage{microtype}

\usepackage{inconsolata}

\usepackage{amsthm,amsmath,amssymb}

\usepackage{mathrsfs}
\usepackage{caption}

\usepackage{dutchcal}

\usepackage{array}

\usepackage{colortbl}
\usepackage{float}
\usepackage{algorithm}
\usepackage{algpseudocode}
\usepackage{xspace}
\usepackage{arydshln}
\usepackage{makecell}
\usepackage{amsmath} 
\usepackage{mathrsfs} 
\usepackage{subfig}
\usepackage{hyperref}

\definecolor{LightYellow}{rgb}{1.0, 1.0, 0.96} 
\definecolor{SkyBlue}{rgb}{0.98, 0.98, 1.0}
\definecolor{LightGreen}{rgb}{0.88, 1.0, 0.88}
\definecolor{LightPink}{rgb}{1.0, 0.98, 0.98}
\definecolor{DarkYellow}{RGB}{255,215,0} 
\definecolor{DarkGreen}{RGB}{0,100,0}
\definecolor{DarkBlue}{RGB}{0,0,139}
\definecolor{Brown}{RGB}{165,42,42}

\newcolumntype{y}{>{\columncolor{LightYellow}}c}
\newcolumntype{b}{>{\columncolor{SkyBlue}}c}
\newcolumntype{g}{>{\columncolor{LightGreen}}c}
\newcolumntype{k}{>{\columncolor{LightPink}}c}
\newcolumntype{P}[1]{>{\centering\arraybackslash}p{#1}}

\usepackage{graphicx} 
\usepackage{epstopdf}

\newcommand{\method}{\textsc{ReClaim}\xspace}

\title{Ground Every Sentence: Improving Retrieval-Augmented LLMs with Interleaved Reference-Claim Generation}

\author{
    Sirui Xia$^{1}$, Xintao Wang$^{1}$, Jiaqing Liang$^{2*}$, Yifei Zhang$^{1}$, Weikang Zhou$^{3}$ \\
    \textbf{Jiaji Deng$^{3}$, Fei Yu$^{3}$, Yanghua Xiao$^{1*}$} \\ 
    $^1$Shanghai Key Laboratory of Data Science, School of Computer Science, Fudan University \\
    $^2$School of Data Science, Fudan University \quad
    $^3$AntGroup \quad \\
    \texttt{\{srxia24, xtwang21, yifeizhang23\}@m.fudan.edu.cn,} \\
    \texttt{\{liangjiaqing, shawyh\}@fudan.edu.cn,} \\ 
    \texttt{feiyu.fyyu@gmail.com},\ \ \ \texttt{\{zhouweikang.zwk, dengjiaji.djj\}@antgroup.com} 
}

\begin{document}
\maketitle
\begin{abstract}
Retrieval-Augmented Generation (RAG) has been widely adopted to enhance Large Language Models (LLMs) in knowledge-intensive tasks. 
To enhance credibility and verifiability in RAG systems, Attributed Text Generation (ATG) is proposed, which provides citations to retrieval knowledge in LLM-generated responses. 
Prior methods mainly adopt coarse-grained attributions, with passage-level or paragraph-level references or citations, which fall short in verifiability. 
This paper proposes \method~(\textbf{Re}fer \& \textbf{Claim}), a fine-grained ATG method that alternates the generation of references and answers step by step. 
Different from previous coarse-grained attribution, \method~provides sentence-level citations in long-form question-answering tasks. 
With extensive experiments, we verify the effectiveness of \method in extensive settings, achieving a citation accuracy rate of 90\%.\footnote{Code and datasets are public at: \url{https://github.com/pdxthree/ReClaim}}
\end{abstract}

\section{Introduction}

Retrieval-Augmented Generation (RAG)~\cite{lewis2020retrieval} is a technique that integrates information retrieval with natural language generation to enhance the performance of large language model (LLMs) responses. However, the RAG system still encounters challenges related to verifiability and credibility. To address these issues, attributed text generation (ATG) \cite{bohnet2022attributed} has been proposed. ATG aims to improve RAG systems in terms of: 1) Credibility, as explicit citations can reduce hallucinations; 2) Verifiability, making it easier for users to verify the answer.

Previous efforts on ATG mainly focus on passage-level \citep{thoppilan2022lamda} or paragraph-level references \citep{menick2022teaching,nakano2021webgpt,gao2023enabling}. Although these attribution methods have contributed to improving the verifiability and credibility of model responses, current methods often focus on relatively coarse-grained attributions, which may contain a significant amount of irrelevant information. This increases the time required for fact-checking.

\begin{figure}[t]
    \centering
    \includegraphics[width=0.48\textwidth]{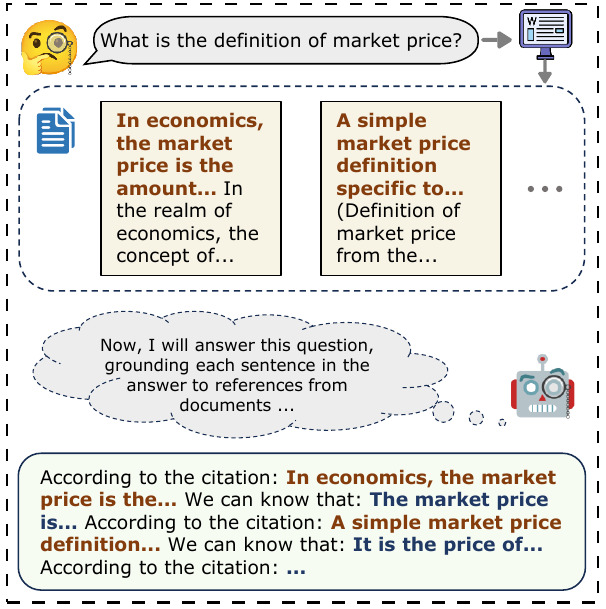}
    \caption{The task setup for \method. Given question and reference passages from a large corpus. The LLM then generates a response with fine-grained citations. For detailed examples, see Table~\ref{table:testdatacase}.}
    \label{fig:picture1}
\end{figure}

In this paper, we propose \method, which generates attributed text with interleaving references and answers for RAG systems, as is shown in Figure~\ref{fig:picture1}. 
This method enables sentence-level fine-grained attributions in long-form question-answering using the RAG system.

To enhance the LLM's performance in attributed text generation, we developed a training dataset and fine-tuned the LLM to facilitate sentence-level citation selection from given reference passages and subsequent answer generation. We implemented an alternating strategy between citation generation and answer text generation. We apply constrained decoding during LLM inference by encoding reference passages into a token-level prefix tree. This restricts the LLM to generate citations only along the tree's paths, ensuring alignment with the reference passages and avoiding inconsistencies.

The results of experiments demonstrate that \method outperforms existing baselines. \method significantly improves the citation quality, enabling the citations to better support the answer. Furthermore, \method greatly reduces the verbosity of citations, thereby easing the fact-checking process. 

Our contributions are summarized as follows:
\begin{enumerate}
    \item We propose a method called \method, which alternately generates citations and answer sentences, enabling LLM to produce answers with sentence-level citations, thus enhancing the LLM's verifiability and credibility.
    \item To enhance LLMs in sentence-level citation generation, we construct a dataset based on WebGLM-QA~\cite{liu2023webglm} and ELI5~\cite{fan2019eli5} dataset. Then, we fine-tune Llama3-8B-Instruct~\cite{dubey2024llama} models for reference and claim generation respectively, achieving better citation quality compared to the baseline method with ChatGPT~\cite{chatgpt}.
    \item Through extensive experiments, we demonstrate the effectiveness of our method in enhancing the LLM's verifiability and credibility, achieving performance comparable to much bigger models like ChatGPT.
\end{enumerate}

\section{Related Work}

\paragraph{Retrieval-Augmented Generation}
In this paper, we use the RAG (Retrieval-Augmented Generation) system to generate answer with citations. The RAG system was proposed to combine information retrieval with generation models for tasks such as question answering and knowledge generation. Similarly, this system has been widely applied to handle complex tasks that require extracting information from a large number of documents, including open-domain question answering, dialogue systems, and information summarization~\cite{izacard2021leveraging,karpukhin2020dense}.

\paragraph{Long-form Text Question Answering} Our work primarily focuses on the long-form question answering (LFQA) task within the RAG system. Unlike short-form QA \cite{rajpurkar2016squad,joshi2017triviaqa}, which concentrates on extracting concise facts, LFQA generates comprehensive answers that require deep contextual understanding and information integration from multiple sources~\cite{fan2019eli5,stelmakh2022asqa,malaviya2024expertqa}.

\paragraph{Attributed Text Generation}
Many current works propose various methods for generating answer text with citations, differing in their approaches to attribution and citation granularity.

LaMDA \cite{thoppilan2022lamda} provides attribution for the entire response in the form of URLs pointing to entire documents. WebGPT \cite{nakano2021webgpt} and GopherCite \cite{menick2022teaching} use reinforcement learning from human preferences to enable LLMs to answer questions while providing snippet citations. ALCE \cite{gao2023enabling} goes further by providing one or more input documents as attribution for each sentence in the answer, in a manner similar to cross-referencing. 
Additionally, some work has focused on fine-tuning models to improve the generation of attributed answer text~\cite{huang2024training, asai2023self, huang2024learning}. 

In addition to the aforementioned methods that add citations directly during answer generation, there are some works that focus on finding citations afterward~\cite{gao2023rarr}. Some research further achieves better attribution performance through multiple retrievals and validations~\cite{li2024llatrieval,sun2023towards}.

\section{Method}
We introduce \method, which aims to generate text with sentence-level citations. 
Specifically, \method generates citations with answers alternatively, in a step-by-step manner.  
We first introduce the overview of \method~in Section~\ref{sec:ReClaim}, and then detail the  specific implementation in Section~\ref{sec:dataset} to~\ref{sec:constraint}.

\begin{figure*}[hbt]
    \centering
    \includegraphics[width=\textwidth]{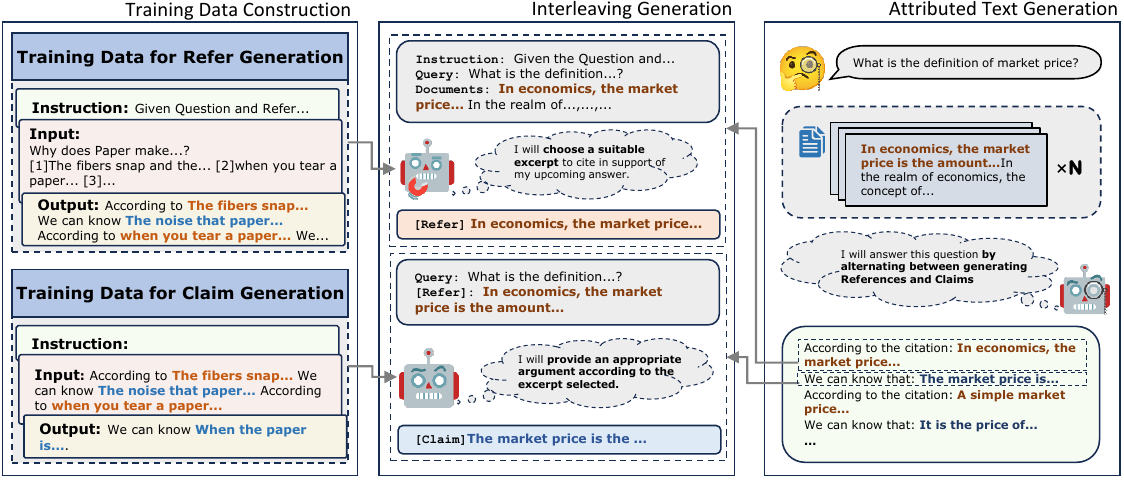}
    \caption{The generation process of \method~w/IG. Based on the given questions and the reference passages retrieved, the LLM alternately generates the reference parts and the claim parts in a step-by-step manner. For these two stages of generation, distinct datasets are constructed to train the base model, which alternately switches between the fine-tuned models and the input context during inference.}
    \label{fig:picture3}
\end{figure*}

\subsection{\method: Interleaving Reference and Claim}
\label{sec:ReClaim}
Our task can be formally expressed as follows: Given a query \textit{q} and several reference passages retrieved by the RAG system $\mathcal{D}$, the LLM is required to generate an output $\mathcal{O}$ = \{ $\textit{r}_\textit{1}$, $\textit{c}_\textit{1}$, $\textit{r}_\textit{2}$, $\textit{c}_\textit{2}$...$\textit{r}_\textit{n}$, $\textit{c}_\textit{n}$ \}, where $\mathcal{O}$ consists of n sentence-level fine-grained references $\textit{r}_\textit{1}$, ..., $\textit{r}_\textit{n}$, which represent the fine-grained citations coming from reference passages (provided as text, not just source numbers) and n claims $\textit{c}_\textit{1}$, ..., $\textit{c}_\textit{n}$, which are portions of the LLM's response generated based on these references. 
Each reference $\textit{r}_\textit{i}$ serves as a substantiation of claim $\textit{c}_\textit{i}$, and together, all $\textit{c}_\textit{i}$ form the LLM's complete answer to the question.

During generation, the LLM alternates between generating references and claims. However, experiments revealed that the LLM faces several issues: 1) The generated references are not always consistent with the retrieved passages; 2) The generated claims do not always attribute well to the corresponding references. Therefore, in the following sections, we study how to improve the generation of references and claims.

\subsection{Training Dataset Construction}
\label{sec:dataset}

To improve the LLM's ability to choose the references and generate corresponding claims, we construct a specialized fine-tuning dataset based on the WebGLM-QA~\cite{liu2023webglm} and ELI5~\cite{fan2019eli5} datasets. 

The training data input includes an instruction, a query, and reference passages. The output consists of multiple reference-claim pairs, with the reference being a sentence-level citation and the claim being a part of the answer. The reference and claim components are designed to train the LLM’s ability to select accurate and relevant fine-grained citations from the complete reference passages, and to generate fluent, coherent answer texts that are faithful to the selected citations.

The steps of our dataset construction are detailed as follows: 

\paragraph{Reference Passages Retrieval}
In the WebGLM-QA Dataset, the reference passages are fine-grained texts closely aligned with the query. However, some research has shown that irrelevant reference texts significantly impact the quality of LLM-generated answers~\cite{wu2024how}. To enhance the LLM's performance against irrelevant contexts, we sample a portion of the constructed WebGLM-QA training dataset, retrieve the top-100 passages using BM25, and calculate the similarity between the query and passages with a Reranker model~\cite{chen2024bge}. Passages with high BM25 ranking but low re-rank score are selected as irrelevant text and added to the training data.

For the ELI5 dataset, we use BM25 to retrieve the top-100 reference passages, which included some irrelevant noise to simulate real-world retrieval conditions. Additionally, we use the Reranker to re-rank the top 100 passages retrieved from a subset of the selected ELI5 dataset, producing a refined re-ranked dataset.

\paragraph{Model Answer Generation}
For the WebGLM-QA dataset, we directly use the original model responses. For the ELI5 datasets, due to significant information discrepancies between the human answers and the retrieved passages, we provide the top-5 retrieved passages to the LLM (Llama-3.1-405B-Instruct~\cite{dubey2024llama}) to generate long-form answers.

\paragraph{Multi-Stage Citation Search}
To select sentence-level fine-grained citations for the answers generated by the LLM, we employ a multi-stage citation search approach that moves from coarse to fine. 

We first enable the Llama-3.1-405B-Instruct model to automatically segment the text into clauses. For each clause, the model identifies the minimal set of citations from the reference passages that sufficiently support it.

Then, we use NLI (Natural Language Inference) model~\cite{honovich2022true} to evaluate the entailment relationship between the chosen citations (as the premise) and the corresponding clause (as the hypothesis). To enhance filtering precision, we set a threshold (\(\theta = 0.8\)) to discard cases where the entailment probability between the chosen citation and the clause falls below the threshold.

\begin{table}[h]
\centering
\resizebox{0.48\textwidth}{!}{%
    \begin{tabular}{ccccccc}
      \toprule
      \multirow{2}{*}{\textbf{Dataset}} & \multirow{2}{*}{\textbf{Samples}} & \multicolumn{3}{c}{\textbf{Average Length}}
      \\
      \cmidrule(lr){3-5}
      & & \multicolumn{1}{c}{\textbf{Answer}} & \multicolumn{1}{c}{\textbf{Citation}} & \multicolumn{1}{c}{\textbf{Passages}} \\
      \midrule
      WebGLM-QA & 4582 & 98.53 & 154.62 & 326.57\\
      WebGLM-QA Extend & 2605 & 83.85 & 114.1 & 396.16\\
      ELI5 Default & 3383 & 91.33 & 132.34 & 545.01\\
      ELI5 Rerank & 2653 & 107.54 & 158.52 & 542.02\\
    
      \bottomrule
    \end{tabular}
}
\caption{Statistics of the training dataset. For more details, please refer to the appendix~\ref{sec:detail_dataset}.}
\label{table:traindataset}
\end{table}

\subsection{Unified Generation}
\label{sec:unified}

The \method$_\texttt{Unified}$ method uses a simple fine-tuning and inference approach. It first performs instruction fine-tuning on the LLM using the dataset constructed in Section~\ref{sec:dataset}. Then, it uses the fine-tuned LLM to perform one-step generation. Based on the given query and reference passages, it directly outputs the attributed answer. This generation process can be described as: $UnifiedGen = \{\{\textit{r}_\textit{1}, \textit{c}_\textit{1}, ..., \textit{r}_\textit{i}, \textit{c}_\textit{i}\} \ |\  Query, Passages\}$.

\subsection{Interleaving Generation}
\label{sec:interleave}

During the claim generation stage, since the LLM has already selected sufficiently granular reference text to follow, which contain the answer information, the full input context is not required. 
Therefore, the \method~w/IG method trains separate LLMs for the generation of the reference parts and the claim parts, and alternates between the two LLMs during answer generation, adjusting the input to each LLM accordingly (IG represents the interleaving use of two fine-tuned LLMs for the iterative generation of references and claims).

The entire generation process, as illustrated in Figure~\ref{fig:picture3}, involves the following steps to train the LLMs and generate the answer.

\paragraph{Reference Generation} 
\label{sec:refgeneration}
During the generation of reference parts, the LLM needs to generate the next reference based on the complete input context and previous output. We define this generation process as $ReferGen = \{\textit{r}_\textit{i+1} \ |\  Prompt, \{\textit{r}_\textit{1}, \textit{c}_\textit{1}, ..., \textit{r}_\textit{i}, \textit{c}_\textit{i}\}\}$, where $\textit{r}_\textit{i+1}$ refers to the reference part generated in the next stage, $Prompt$ refers to the complete input context containing instructions, query and reference passages, and $\textit{r}_\textit{1}, \textit{c}_\textit{1}, ..., \textit{r}_\textit{i}, \textit{c}_\textit{i}$ refer to the previously generated references and claims. 
As the training of the LLM for generating the reference parts does not require masking parts of the input context information, we follow the same approach as in the Section~\ref{sec:unified} to fine-tune the ReferModel.

\paragraph{Claim Generation}
\label{sec:claimgeneration}

During the generation of claim parts, the LLM only needs to generate the next claim based on the previous output. We define this generation process as $ClaimGen = \{\textit{c}_\textit{i+1} \ | \{\textit{r}_\textit{1}, \textit{c}_\textit{1}, ..., \textit{r}_\textit{i}, \textit{c}_\textit{i}, \textit{r}_\textit{i+1}\}\}$, where $\textit{c}_\textit{i+1}$ refers to the claim part generated in the next stage, and $\textit{r}_\textit{1}, \textit{c}_\textit{1}, ..., \textit{r}_\textit{i}, \textit{c}_\textit{i}, \textit{r}_\textit{i+1}$ refer to the previously generated references and claims. 
We utilize the training dataset from Section~\ref{sec:dataset}, formatting it to align with our ClaimGen generation process, and then use this formatted dataset to fine-tune the ClaimModel.\\


\subsection{Decoding Constraints}
\label{sec:constraint}

To ensure the generated reference parts align with the retrieved reference passages, we apply decoding constraints through the following three steps:

\paragraph{Sentence Segmentation and Encoding} 
We segment the reference passages into individual sentences. Then, we use the LLM tokenizer to encode these sentences into vectors. Each vector representation of a sentence serves as the smallest unit for generating the reference parts. 

\paragraph{Constructing Prefix Tree} 
The encoded vectors are transformed into a list format and organized into a Prefix tree~\cite{fredkin1960trie} structure. Employing such a structure to store our reference sentences facilitates the choice of the next token in subsequent generation steps.

\paragraph{Constrained Inference} 
During the inference stage for generating reference parts, we select the token with the highest generation probability that satisfies the current prefix tree path as the next output token. This process continues until reaching a leaf node. Upon reaching a leaf node, the LLM either select another prefix tree path for output or begin the claim generation. 
This approach allows us to select a complete and consistent sentence from the reference passages as part of our current reference each time.

\section{Experimental Protocol}
In this section, we employ the GPT models and several open-source LLMs to validate the effectiveness of our method across multiple evaluation dimensions. We conduct a comprehensive analysis by assessing the performance of our approach on various metrics. 

\subsection{Evaluation Datasets}
\paragraph{ASQA} We evaluate the 948 samples from the ASQA dataset~\cite{stelmakh2022asqa}, selected by ALCE, and use the five oracle paragraphs provided by ALCE as reference passages, which are chosen from the top 100 retrieved passages representing the gold passages.
\paragraph{ELI5} We evaluate the 1,000 samples selected from ELI5 dataset~\cite{fan2019eli5} by ALCE and use the five oracle passages as reference passages.
\paragraph{EXPERTQA} To test the LLM's generalization ability under the standard RAG process, we select 1,000 samples from the EXPRTQA~\cite{malaviya2024expertqa} dataset. We follow the standard RAG procedure: using BM25 to retrieve the top-100 passages, then re-ranking them to select the top 5 passages.

\subsection{Evaluation Metrics}
\label{sec:metrics}
Building on our previous task definition, we focus on evaluating the LLM-generated outputs in three key areas. Below, we introduce three evaluation dimensions, with more detailed information and calculation methods provided in Appendix~\ref{sec:detail_metric}.

\paragraph{Answer Quality}
We concatenate all claim parts in order to form the answer to the question, and follow the ALCE evaluation method to calculate the correctness and fluency of the answers. 
For answer correctness, in the ASQA dataset, we use Exact Match Recall (EM Rec.) to measure the percentage of golden short answers contained in the generated answers. In the ELI5 and EXPERTQA dataset, we adopt Claim Recall (Claim Rec.) to measure the percentage of key claims included in the answers. 
To evaluate answer fluency, we use MAUVE \cite{pillutla2021mauve} to measure the similarity between output and gold answer.

\paragraph{Citation Quality}
Similar to ALCE, we employ the AutoAIS \cite{bohnet2022attributed} to measure the citation quality. Our citation quality is also measured by two metrics: 1) Correct Attribution Score (CAS), which determines whether the answer is entirely supported by cited sentences and is the most critical metric in our evaluation; 2) Citation Redundancy Score (CRS), which identifies any redundant citation sentences.

We use the NLI~\cite{honovich2022true} model to compute the entailment relationship between each reference part and the corresponding claim part, and the final CAS score is the proportion of correctly attributed sentences in the answer.

We also use the NLI model need to determine if the reference contains redundant sentences, and the final CRS score is the proportion of non-redundant sentences relative to all sentences in the references.

\paragraph{Verifiability}
We employ three metrics to measure the verifiability:
1) Citation length, where shorter citation text typically reduces the time needed for fact-checking;
2) Attribution Ratio (AR), which represents the proportion of sentences in the output that are attributed;
3) Consistency Ratio (CR), which represents the proportion of text consistency between the reference parts and the reference passages through string matching.

\setlength\tabcolsep{2.8pt}
\begin{table*}[ht]
\centering
\resizebox{\textwidth}{!}{%
    \begin{tabular}{llyyyybbbbkkkk}
      \toprule
      \multirow{3}{*}{\textbf{Method}} & \multirow{3}{*}{\textbf{Model}} & \multicolumn{4}{c}{\textbf{ASQA}} & \multicolumn{4}{c}{\textbf{ELI5}} & \multicolumn{4}{c}{\textbf{EXPERTQA}} \\
      \cmidrule(lr){3-6}\cmidrule(lr){7-10}\cmidrule(lr){11-14}
       & & \multicolumn{1}{c}{\textbf{Fluency}} & \multicolumn{1}{c}{\textbf{Correct.}} & \multicolumn{2}{c}{\textbf{Citation}} & \multicolumn{1}{c}{\textbf{Fluency}} & \multicolumn{1}{c}{\textbf{Correct.}} & \multicolumn{2}{c}{\textbf{Citation}} &
       \multicolumn{1}{c}{\textbf{Fluency}} & \multicolumn{1}{c}{\textbf{Correct.}} & \multicolumn{2}{c}{\textbf{Citation}} \\
       \cmidrule(lr){3-3}\cmidrule(lr){4-4}\cmidrule(lr){5-6}\cmidrule(lr){7-7}\cmidrule(lr){8-8}\cmidrule(lr){9-10}\cmidrule(lr){11-11}\cmidrule(lr){12-12}\cmidrule(lr){13-14}
       & & \multicolumn{1}{c}{\textbf{\small MAUVE}} & \multicolumn{1}{c}{\textbf{\small EM Rec.}} & \multicolumn{1}{c}{\textbf{\small CAS}} & \multicolumn{1}{c}{\textbf{\small CRS}} & \multicolumn{1}{c}{\textbf{\small MAUVE}} & \multicolumn{1}{c}{\textbf{\small Claim Rec.}} & \multicolumn{1}{c}{\textbf{\small CAS}} & \multicolumn{1}{c}{\textbf{\small CRS}} & \multicolumn{1}{c}{\textbf{\small MAUVE}} & \multicolumn{1}{c}{\textbf{\small Claim Rec.}} & \multicolumn{1}{c}{\textbf{\small CAS}} & \multicolumn{1}{c}{\textbf{\small CRS}} \\
      \midrule
      \rowcolor{gray!30}
      \multicolumn{14}{c}{\textit{Prompting-Based}} \\
      \multirow{3}{*}{{ALCE}} 
       & ChatGPT & 64.4 & 48.9 & 74.5 & 72.7 & 59.4 & \textbf{21.3} & 57.8 & 56.0 & 53.7 & 20.5 & 66.7 & 64.9 \\
       & Llama3-8B & 79.2 & 55.2 & 54.7 & 54.6 & 45.4 & 20.5 & 42.8 & 39.3 & 56.7 & 20.2 & 51.4 & 47.8 \\
       & \quad w/Rerank & 81.3 & 55.3 & 79.1 & 76.4 & 37.6 & 20.0 & 59.5 & 53.8 & 45.3 & 19.8 & 68.7 & 60.2\\
       \rowcolor{gray!30}
       \multicolumn{14}{c}{\textit{Post-hoc}} \\ 
       ClosedBook & Llama3-8B & 50.5 & 28.9 & 13.7 & 13.7 & 66.8 & 16.8 & 17.4 & 17.4 & 37.6 & \textbf{25.4} & 11.2 & 11.2 \\
       Open-book & Llama3-8B & 73.7 & 53.2 & 52.1 & 52.1 & 26.6 & 20.8 & 31.9 & 32.0 & 58.7 & 21.0 & 36.9 & 36.8 \\
       
       \rowcolor{gray!30}
       \multicolumn{14}{c}{\textit{Training-Based}} \\ 
        Self-RAG & Llama2-7B & 70.6 & 38.7 & 53.3 & 66.2 & 33.1 & 9.7 & 23.1 & 33.9 & 19.0 & 12.5 & 9.4 & 12.1 \\
        RS+RL & Llama2-7B & 84.4 & 47.7 & 75.5 & 69.4 & 43.6 & 19.1 & 59.1 & 51.6 & 46.6 & 15.3 & 67.8 & 60.1 \\
        FRONT & Llama2-7B & 72.5 & 56.5 & 72.2 & 66.0 & 49.7 & 18.1 & 64.0 & 59.1 & 56.5 & 14.7 & 73.6 & \textbf{68.9} \\
       
       \rowcolor{gray!30}
       \multicolumn{14}{c}{\textit{Our Methods}} \\ 
       \multirow{1}{*}{{0-shot}} 
       & GPT-4o & 72.9 & 52.8 & 74.8 & 51.6 & 37.3 & 19.9 & 63.5 & 30.7 & 59.4 & 18.5 & 73.4 & 26.5 \\
       
       \hdashline
       \multirow{3}{*}{{3-shot}} 
       & Llama3-8B & 90.1 & 50.7 & 77.7 & 62.1 & 61.3 & 17.9 & 78.3 & 45.3 & 46.0 & 12.6 & 79.8 & 51.4 \\
       & ChatGPT & 74.9 & 52.6 & 72.5 & 63.4 & 27.8 & 17.8 & 68.6 & 50.8 & 30.6 & 14.1 & 72.7 & 55.7\\
       & GPT-4o & \textbf{91.3} & \textbf{56.6} & 77.4 & 58.0 & 29.7 & 21.1 & 70.2 & 36.8 & 38.7 & 18.2 & 68.0 & 45.2 \\

       \hdashline
       \multirow{1}{*}{{\small \method$_\texttt{Unified}$}}
       & Llama3-8B & 89.8 & 53.3 & 68.2 & 58.9 & \textbf{73.6} & 19.9 & 69.4 & 48.6 & 64.2 & 16.4 & 70.0 & 50.4 \\

       \hdashline
       \multirow{2}{*}{{\small \method~w/IG}} 
       & Llama2-7B & 71.4 & 55.0 & 89.5 & 78.7 & 67.3 & 17.6 & 86.3 & 58.6 & \textbf{67.2} & 12.7 & 86.3 & 60.2 \\
       & Llama3-8B & 88.1 & 53.5 & \textbf{92.1} & \textbf{86.1} & 71.6 & 17.8 & \textbf{89.9} & \textbf{67.5} & 63.5 & 14.0 & \textbf{90.1} & 68.6 \\
       
      \bottomrule
    \end{tabular}
}
\caption{Results on ASQA, ELI5, EXPERTQA. Definitions for Fluency, Correct. and Citation are in Section~\ref{sec:metrics}.}
\label{table:main}
\end{table*}

\subsection{Baselines}
We use only methods that restrict citation sources to the top-k reference passages as our comparison baselines. We exclude methods that introduce additional retrieval steps, such as VTG~\cite{sun2023towards} and LLatrieval~\cite{li2024llatrieval}, from our comparison.

\paragraph{Prompting-Based} Following the ALCE~\cite{gao2023enabling} method, we prompt ChatGPT and Llama3.1-8B-Instruct with few-shot demonstrations that consist of a query, the top-5 retrieved passages, and an answer with inline citations. Additionally, we perform four samples of the generation from Llama3.1-8B-Instruct and select the best answer from each case as the experimental result for the w/Rerank method.

\paragraph{Post-hoc}
Following the ALCE method, we generate answer using the Llama3-8B-Instruct model with the closed-book approach based on the given query. Then, for each statement in the answer, we search the top-100 passages for citations.

We also provide the LLMs with the top-5 reference passages, allowing it to generate the answer based on them and then re-identify citations within the top-5 passages. We refer to this method as open-book.

\paragraph{Training-Based}
Following Self-RAG~\cite{asai2023self}, the LLM is trained to retrieve passages on demand, verify relevance, and generate answers based on the retrieved content. We directly provide the selfrag-7B (fine-tuned Llama2-7B model) with the top-5 passages for generation.

Following the method in~\cite{huang2024training}, we use the LLM trained with fine-grained rewards (RS+RL, fine-tuned Llama2-7B model) to generate answer with citations. 
Additionally, we compare the fine-grained attribution method FRONT~\cite{huang-etal-2024-learning}, which is fine-tuned based on the Llama2-7B model.

\subsection{Methods}
We evaluate \method by testing several LLMs with different generation approaches.

\paragraph{\method~prompting}
We directly prompt LLMs to generate answer with citations, using GPT models \cite{openai2023gpt4,chatgpt} and the Llama series \cite{touvron2023llama} to evaluate the performance and effectiveness.

\paragraph{\method$_\texttt{Unified}$}
We follow the methodology in Section~\ref{sec:unified} and use the training dataset constructed in Section~\ref{sec:dataset} to fully fine-tune the Llama3-8B-Instruct model, then conduct one-step generation with citations to evaluate the effectiveness of the \method~$_\texttt{Unified}$ method.

\paragraph{\method~w/IG}
As outlined in Section~\ref{sec:interleave}, we fine-tune the same base model (Llama2-7B-hf and Llama3-8B-Instruct), generating two separate LLMs. During the inference phase, we alternate between these two LLMs to interleavingly generate the reference and claim parts. \\

For these two fine-tuning methods, we adopt the constrained decoding described in Section~\ref{sec:constraint} to limit the generation of reference parts. For more experimental details, see Appendix~\ref{sec:detail_experiment}.

\setlength\tabcolsep{2.8pt}
\begin{table*}[h]
\centering
\small
\resizebox{\textwidth}{!}{%
    \begin{tabular}{lyyyybbbbkkkk}
    
      \toprule
      \multirow{3}{*}{\textbf{Method}} & \multicolumn{4}{c}{\textbf{ASQA}} & \multicolumn{4}{c}{\textbf{ELI5}} & \multicolumn{4}{c}{\textbf{EXPERTQA}} \\
      \cmidrule(lr){2-5}\cmidrule(lr){6-9}\cmidrule(lr){9-13}
       & \multicolumn{2}{c}{\textbf{Length}} & \multicolumn{1}{c}{\textbf{Consistency}} & \multicolumn{1}{c}{\textbf{Attri.}} 
       & \multicolumn{2}{c}{\textbf{Length}} & \multicolumn{1}{c}{\textbf{Consistency}} & \multicolumn{1}{c}{\textbf{Attri.}} 
       & \multicolumn{2}{c}{\textbf{Length}} & \multicolumn{1}{c}{\textbf{Consistency}} & \multicolumn{1}{c}{\textbf{Attri.}} \\
       \cmidrule(lr){2-3}\cmidrule(lr){4-4}\cmidrule(lr){5-5}\cmidrule(lr){6-7}\cmidrule(lr){8-8}\cmidrule(lr){9-9}\cmidrule(lr){10-11}\cmidrule(lr){12-12}\cmidrule(lr){13-13}
       & \multicolumn{1}{c}{\textbf{Citation}} & \multicolumn{1}{c}{\textbf{Claim}} & \multicolumn{1}{c}{\textbf{CR}} & \multicolumn{1}{c}{\textbf{AR}} & \multicolumn{1}{c}{\textbf{Citation}} & \multicolumn{1}{c}{\textbf{Claim}} & \multicolumn{1}{c}{\textbf{CR}} & \multicolumn{1}{c}{\textbf{AR}} & \multicolumn{1}{c}{\textbf{Citation}} & \multicolumn{1}{c}{\textbf{Claim}} & \multicolumn{1}{c}{\textbf{CR}} & \multicolumn{1}{c}{\textbf{AR}} \\
      \midrule
      
       ALCE & 536.3 & 85.5 & 100.0 & 91.3 & 660.0 & 98.09 & 100.0 & 96.9 & 627.5 & 115.1 & 100.0 & 84.1 \\
       RS+RL & 327.0 & 39.9 & 100.0 & 94.5 & 476.6 & 80.8 & 100.0 & 93.2 & 501.9 & 81.9 & 100.0 & 95.1\\
       \midrule
       \method$_\texttt{3-shot}$ & 106.8 & 59.8 & 75.5 & 100.0 & 162.7 & 82.1 & 72.1 & 100.0 & 198.7 & 82.1 & 77.5 & 100.0 \\
       \method$_\texttt{Unified}$ & 77.9 & 52.9 & 100.0 & 100.0 & 139.8 & 93.1 & 100.0 & 100.0 & 150.9 & 99.2 & 100.0 & 100.0\\
       \method~w/IG & 82.8 & 68.9 & 100.0 & 100.0 & 145.0 & 104.7 & 100.0 & 100.0 & 157.5 & 109.1 & 100.0 & 100.0\\     
      \bottomrule
    \end{tabular}
}
\caption{The generated text length, consistency of references, and proportion of attributed answer sentences in different methods. \method$_\texttt{3-shot}$ does not use decoding constraints, while \method$_\texttt{Unified}$ and \method~w/IG use decoding constraints.}
\label{table:secondary}
\end{table*}

\setlength\tabcolsep{2.8pt}
\begin{table*}[h]
\centering
\resizebox{\textwidth}{!}{%
    \begin{tabular}{lyyyybbbbkkkk}
      \toprule
      \multirow{2}{*}{\textbf{Method}} & \multicolumn{4}{c}{\textbf{ASQA}} & \multicolumn{4}{c}{\textbf{ELI5}} & \multicolumn{4}{c}{\textbf{EXPERTQA}} \\
      \cmidrule(lr){2-5}\cmidrule(lr){6-9}\cmidrule(lr){10-13}
       
       & \multicolumn{1}{c}{\textbf{\small MAUVE}} & \multicolumn{1}{c}{\textbf{\small EM Rec.}} & \multicolumn{1}{c}{\textbf{\small CAS}} & \multicolumn{1}{c}{\textbf{\small CRS}} & \multicolumn{1}{c}{\textbf{\small MAUVE}} & \multicolumn{1}{c}{\textbf{\small Claim Rec.}} & \multicolumn{1}{c}{\textbf{\small CAS}} & \multicolumn{1}{c}{\textbf{\small CRS}} & \multicolumn{1}{c}{\textbf{\small MAUVE}} & \multicolumn{1}{c}{\textbf{\small Claim Rec.}} & \multicolumn{1}{c}{\textbf{\small CAS}} & \multicolumn{1}{c}{\textbf{\small CRS}} \\
      \midrule
       \multirow{1}{*}{{ReferModel-Only w/Extend}}
       & 29.4 & 29.8 & 70.5 & 58.9 & \underline{61.4} & 13.4 & 59.9 & 40.6 & \underline{50.0} & 10.1 & 61.7 & 44.4 \\
       \multirow{1}{*}{{ReferModel-Only w/Sum}}
       & 29.2 & 33.0 & \textbf{94.5} & \underline{80.8} & 46.8 & 14.6 & \textbf{96.2} & \underline{61.4} & 49.5 & 11.0 & \textbf{97.7} & \textbf{69.1} \\
       \multirow{1}{*}{{ClaimModel-Only}}
       & \underline{43.6} & \textbf{57.7} & 89.6 & 77.2 & 54.5 & \textbf{18.9} & 84.4 & 55.1 & 43.2 & \textbf{14.8} & 85.6 & 57.2 \\
       \multirow{1}{*}{{\method~w/IG}} 
       & \textbf{88.1} & \underline{53.5} & \underline{92.1} & \textbf{86.1} & \textbf{71.6} & \underline{17.8} & \underline{89.9} & \textbf{67.5} & \textbf{63.5} & \underline{14.0} & \underline{90.1} & \underline{68.6} \\
       
      \bottomrule
    \end{tabular}
}
\caption{Ablation study results. The underline indicates the second largest value.}
\label{table:ablation}
\end{table*}

\subsection{Ablation Study}
We primarily conduct ablation experiments on the \method~w/IG method to investigate the necessity of training data filtering and fine-tuning the ReferModel and ClaimModel. All fine-tuned models are based on the Llama3-8B-Instruct model.

\begin{figure}[ht]
    \centering
    \includegraphics[width=0.48\textwidth]{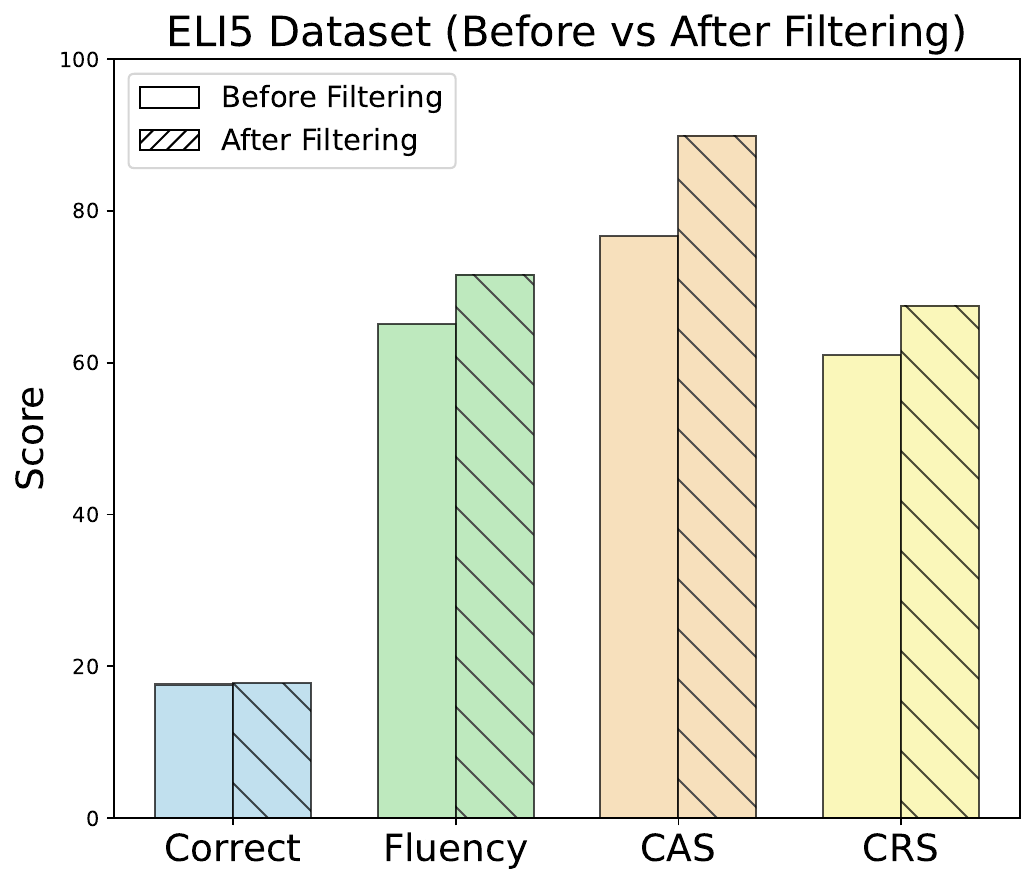}
    \caption{Comparison of the performance of LLMs trained on the training dataset before and after filtering on the ELI5 dataset. The comparative experimental results of the ASQA and EXPERTQA datasets are presented in Appendix Figure~\ref{fig:data_filter_asqa} and Figure~\ref{fig:data_filter_expertqa}.}
    \label{fig:data_filter_eli5}
\end{figure}

\paragraph{Pre-Filtered Training Data} To investigate the necessity of using an NLI model for training data filtering, we fine-tune the LLM on the pre-filtered data while keeping the dataset size and training parameters unchanged. The experimental results are shown in Figure~\ref{fig:data_filter_eli5}.

The experimental results show that fine-tuning the LLM with data filtered by the NLI model improves its performance across various evaluation metrics.

\paragraph{Fine-tuned ReferModel Only} We use the fine-tuned ReferModel to generate reference parts and the base model to generate claim parts. When generating claims, it is essential to ensure that it adheres to the information in the preceding reference and maintains fluency with the previously generated claims. We explore two prompt strategies:
\begin{enumerate}
    \item \textbf{Extension}: We provide the LLM with the previous reference and all preceding claims, requiring the LLM to extend the claims based on the information in the reference.
    \item \textbf{Summary}: We provide the LLM with a reference so that it can directly generate a brief summary based on the previous reference.
\end{enumerate}

\paragraph{Fine-tuned ClaimModel Only} We employ the 3-shot prompting approach to generate the reference section using the base model and utilize the fine-tuned ClaimModel for generating the claim section.

The experimental results are shown in Table~\ref{table:ablation}.

\subsection{Faithfulness Analysis}
To further validate our approach in improving the credibility of model responses, we use GPT-4o-mini~\cite{openai2023gpt4} as our evaluation model to evaluate the faithfulness metric of model answers against complete reference passages using the assessment method proposed by RAGAS~\cite{es2023ragas}. This helps us determine whether the LLM's generated answers are fully based on the information in the given reference paragraphs.

We prioritize the CAS metric and use faithfulness as a supplementary measure to evaluate credibility improvements. Since faithfulness to a citation implies faithfulness to the original text, a high faithfulness score does not guarantee strong CAS performance, as an answer true to the full document may not align with the selected citation.

We evaluate the faithfulness metric on the test datasets, and the results for the ELI5 dataset are shown in Figure~\ref{fig:faithfulness_eli5}. Results for other datasets are shown in Appendix Figure~\ref{fig:faithfulness_asqa} and~\ref{fig:faithfulness_expertqa}.

\begin{figure}[ht]
    \centering
    \includegraphics[width=0.48\textwidth]{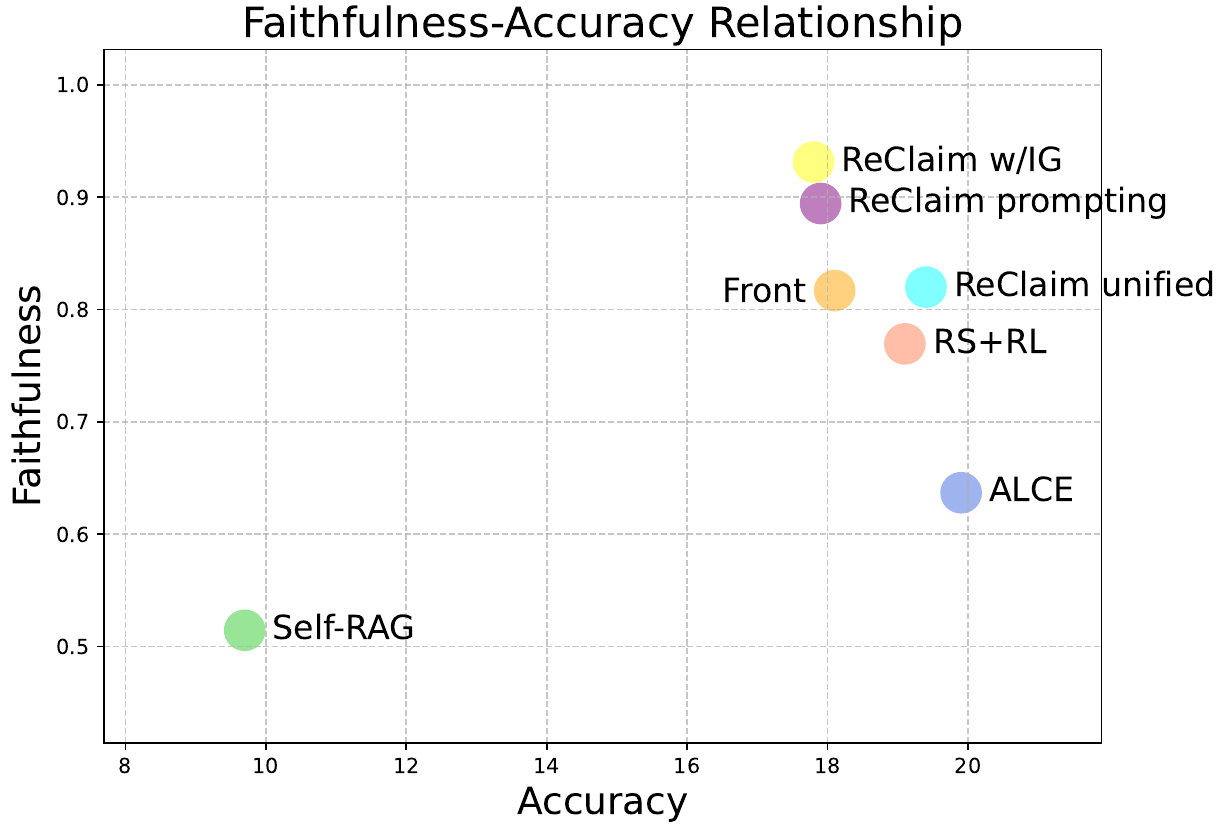}
    \caption{The x-axis represents the accuracy of the LLM's responses, while the y-axis shows the faithfulness score. For the Self-RAG and RS+RL methods, we use the fine-tuned 7B model, whereas for other methods, we employ Llama3-8B-Instruction as the base model.}
    \label{fig:faithfulness_eli5}
\end{figure}

Our method achieved the highest faithfulness score. This demonstrates that while our approach may slightly reduce answer quality, it significantly enhances answer faithfulness and minimizes unnecessary hallucinations linked to the LLM's internal parameters.

The results show an inverse relationship between the accuracy of LLM-generated answers and their faithfulness to reference passages. Higher faithfulness and more granular citations have narrowed the scope of our answer sources, which may contribute to the lower accuracy in LFQA task.

\section{Experiment Results}
In the experiments, we wish to answer two research questions: 
$RQ1)$ How to improve the quality of answers and citations? $RQ2)$ Can \method~enhance the verifiability and credibility of RAG-based question answering?

\subsection{How to Improve the Quality of Answers and Citations?}
The overall performance is presented in Table~\ref{table:main}.

\paragraph{\method~prompting Works}
Experimental results show that directly prompting LLMs yields satisfactory outcomes. Our approach improves average answer fluency and citation accuracy (CAS) compared to baseline methods. Notably, Llama3-8B-Instruct surpasses other baselines in CAS scores, including ALCE+ChatGPT.

Although our method performs worse in CRS, the finer granularity of our citations reduces the impact of redundant content. Redundant citations only add a single irrelevant sentence, which does not significantly increase the cost of fact-checking.

\paragraph{\method$_\texttt{Unified}$ Cannot Improve ATG}
Experimental results show that the \method$_\texttt{Unified}$ method significantly reduces citation quality (CAS). This indicates that fine-tuning the LLM in this way fails to teach it how to generate claims based solely on the information from the previous reference.

\paragraph{\method~w/IG Improves Attribution}
\label{sec:attribution_improve}

Experimental results indicate that the \method~w/IG method outperforms other methods in two citation quality metrics while maintaining high fluency and correctness scores. 

Compared to ALCE using ChatGPT, our method (using Llama3-8B) shows an average improvement of 31.3\% on the CAS, 16.7\% on the CRS, and 25.7\% on the MAUVE across three test datasets, with only a 6.0\% decrease in answer accuracy.

Specifically, we achieved an average CAS score of 90.7 across three test datasets, which is a crucial metric for assessing the degree of text attribution.

Compared to the \method$_\texttt{Unified}$ method, the \method~w/IG approach's biggest difference lies in the training and inference strategies during the claim generation phase. It filters out extraneous contextual information and trains the LLM to generate claims based solely on the preceding reference part. The significant improvements in citation quality demonstrate the effectiveness of the strategy adopted by the \method~w/IG method.

As shown in Table~\ref{table:ablation}, the results of the ablation experiments indicate that fine-tuning two LLMs for alternating generation of references and claims achieves the most balanced performance.

While ReferModel-Only w/Sum method yields a high citation quality score, it compromises the accuracy and fluency of the answers.

On the other hand, using ClaimModel-Only method for generation achieves higher accuracy scores, but it negatively affects answer fluency, and the generation often tends to produce excessively long answers. Additionally, to ensure that the base model adheres to our format when generating reference parts, we need to provide multiple examples, which reduces the effective context window length.

\subsection{Can \method~Enhance the Verifiability and Credibility of RAG-based Question Answering?}

The results of the average length of citations, citation consistency (CR), and attribution ratio (AR) are shown in Table~\ref{table:secondary}.

\paragraph{\method~Improves Verifiability}
Experimental results show that \method's average citation length is about 22\% of the ALCE method, significantly reducing fact-checking time. 

\method ensures that each response sentence is supported by a specific citation source. By employing constrained decoding, our method ensures that every citation is directly extracted from the original text, significantly reducing the occurrence of hallucinations during the citation generation process. Together, these measures collectively enhance the verifiability of the generated answers.

\paragraph{\method~Improve Credibility}
As mentioned in Section~\ref{sec:attribution_improve} regarding the improvement of attribution, this enhancement boosts the credibility of the LLM's responses by providing a definitive source of evidence for the generated answers.

In addition, as shown in Figure~\ref{fig:faithfulness_eli5}, our method increases the faithfulness of the LLM's responses to the overall reference passages by confining the sources of response information within the chosen citations. This improvement reduces the influence of the LLM's internal knowledge on answer generation, thereby enhancing the credibility of our approach.

Our method strictly generates answers based on reference passages, aiming to improve citation quality in LLM responses through stringent constraints. The accuracy of our answers depends heavily on the information density of the reference passages and cannot rely on the LLM's internal knowledge, which may also explain the slight decrease in answer accuracy in our approach.

\section{Conclusion}
We propose a attributed text generation method, \method, which adds sentence-level fine-grained citations to LLM-generated answer in RAG systems. This approach alternates between generating citations and answer sentences, either through prompting or by fine-tuning LLMs.

The results show that our method improves citation quality while maintaining answer quality compared to the baseline methods. Additionally, our approach significantly reduces the length of citations, thus decreasing the time cost required for fact-checking and further enhancing the verifiability of the LLM's responses. Moreover, by using constrained decoding during citation generation, we ensure that each citation is composed of exact sentences from the source passages.

\section*{Limitations}
In this paper, our training dataset was exclusively targeted at long-form question-answering task, which reduces the generalization ability of our fine-tuning methods. 

Additionally, our method requires explicitly outputting the cited sentences, which often leads to generating answers that are more than double the length. This results in increased output time for the LLM.

On one hand, while our approach allows the LLM to synthesize information from multiple reference sentences for attribution, we did not specifically enhance this capability during the training data construction and LLM inference processes. On the other hand, our answer template has certain limitations, as not every response generated by the LLM requires a citation from the reference passages. Therefore, our method has limitations in scenarios that require synthesizing information from multiple sources or necessitate multi-hop reasoning to draw conclusions.

\section*{Ethics Statement}

We hereby acknowledge that all authors of this work are aware of the provided ACL Code of Ethics and honor the code of conduct.

\paragraph{Datasets Source} All original datasets used for training and testing were sourced from open and publicly accessible resources, and they are all approved for use in research purposes, thereby minimizing the risk of sensitive information leakage. 
While we employed LLMs for automated processing during the construction of training dataset, data cleansing was performed to prevent the introduction of additional noise and bias. We solely utilize the constructed dataset for model training.
Although paragraph retrieval is not the focus of our work, the retrieved information from large corpora may introduce noise and bias into LLM-generated responses.
To address these issues, we will optimize the data construction process and explore methods for retrieving noise-free and unbiased information.

\paragraph{AI assistants} AI assistants (ChatGPT) were solely used to improve the grammatical structure of the text.

\bibliography{anthology,custom}
\bibliographystyle{acl_natbib}

\clearpage
\appendix

\section{Notation Table}

\setlength\tabcolsep{3pt}
\begin{table}[h]
\small
  \centering
    \begin{tabular}{P{0.26\columnwidth}|p{0.72\columnwidth}}
    \toprule
    \textbf{} & \textbf{Definition} \\
    \midrule
    \rowcolor[rgb]{ .949,  .953,  .961} \multicolumn{2}{c}{\textit{Task Formulation}} \\

    $\mathcal{q}$ & The given question. \\
    $\mathcal{D}$ & A set of reference passages, where $d\in\mathcal{D}$ is a passage. \\
    $\mathcal{O}$ & The response generated based on the question $\mathcal{q}$ and passages $\mathcal{D}$, composed of $\mathcal{O}= \{\textit{r}_\textit{1}, \textit{c}_\textit{1}, ..., \textit{r}_\textit{i}, \textit{c}_\textit{i}\}$ \\
    $r$ & A portion of the citation, comprising certain sentences from $\mathcal{D}$. \\
    $c$ & A portion of the answer, formed by concatenating all $c$ in sequence as a response to $\mathcal{q}$. \\

    \midrule
    \rowcolor[rgb]{ .949,  .953,  .961} \multicolumn{2}{c}{\textit{Methods}} \\
    ALCE &  A benchmark for Automatic LLMs' Citation Evaluation. \\
    \method-Unified & End-to-end data training and one-step generation. \\
    \method~w/IG & Independent model training and Interleaving Generation. \\
    Citation-only & Use the fine-tuned model to generate references and the base model with 3-shot prompting to generate claims. \\
    Claim-only & Use the base model with 3-shot prompting to generate references and the fine-tuned model to generate claims. \\
    
    \midrule
    \rowcolor[rgb]{ .949,  .953,  .961} \multicolumn{2}{c}{\textit{Metrics}} \\
    \textbf{MAUVE} &Measuring the gap between neural text and human text using divergence frontiers. \\
    \textbf{EM Rec.} & Exact match recall rate of gold short answers in the text generated by the LLM. \\
    \textbf{Claim Rec.} & Recall rate of generated claims in the text generated by the LLM. \\
    \textbf{CAS} & Correct attribution score, the proportion of sentences predicted as correctly attributed among all the sentences in the answer.  \\
    \textbf{CRS} & Citation redundancy score, the proportion of non-redundant sentences relative to all sentences in the references. \\
    \textbf{CR} & Consistency ratio,  the text consistency between the reference parts and the reference passages through string matching.  \\
    \textbf{AR} & Attribution ratio, the proportion of sentences in the output that are attributed. \\
    
    \bottomrule
    \end{tabular}
  \caption{
  The notation table. 
  }
  \label{tab:notations}
\end{table}

In Table~\ref{tab:notations}, we list the notations and abbreviations in this paper, together with their definitions. 

\section{Details of Training Dataset Construction}
\label{sec:detail_dataset}

\subsection{Details of Training Dataset}
\label{statistic_train_data}
The statistics of the training dataset is shown in Table~\ref{table:traindataset}

\paragraph{WebGLM-QA} The WebGLM-QA~\cite{liu2023webglm} dataset is a pioneering resource designed to bolster the development and assessment of web-enhanced question answering systems. It distinguishes itself by seamlessly integrating web search functionalities into the QA process, empowering systems to tap into the expansive repository of knowledge on the internet. Carefully curated to overcome the shortcomings of existing datasets, WebGLM-QA offers a holistic and pragmatic solution for open-domain question answering tasks. It consists of 43,579 high-quality data samples for the train split, 1,000 for the validation split, and 400 for the test split. Refer to our paper for the data construction details.

In WebGLM-QA, the provided answers is generated by the model based on the given reference passages; therefore, we directly use it as our gold answers.

\paragraph{ELI5} ELI5~\cite{fan2019eli5} dataset is a benchmark in natural language processing designed for long-form question answering tasks, focusing on complex and explanatory questions (see Table~\ref{table:testdatacase}). It comprises 270K threads from the Reddit forum “Explain Like I’m Five” (ELI5) where an online community provides answers to questions which are comprehensible by five year olds. For ALCE~\cite{gao2023enabling}, ELI5 questions were paired with passages from Sphere~\cite{piktus2021web}, a filtered version of Common Crawl. 

For the ELI5 dataset, due to the distributional differences between human answers and the retrieved reference passages, we opted to use Llama-3.1-405B-Instruct to generate answers based on the top-5 reference passages as the gold answers.

\subsection{Citation Selection}
During the training data structuring phase, we initially employ Llama-3.1-405B-Instruct to automatically segment the answers within the WebGLM-QA dataset and ELI5 dataset into clauses. Subsequently, for each of these clauses, we undertake a search for corresponding sentence-level, fine-grained citations from the provided reference passages. The prompt we use is as follows:

Upon obtaining answer clauses and their respective citations, structure the response in a cot format, as illustrated in Table~\ref{table:traindatacase}.

\begin{figure}[H]
    \centering
    \includegraphics[width=0.48\textwidth]{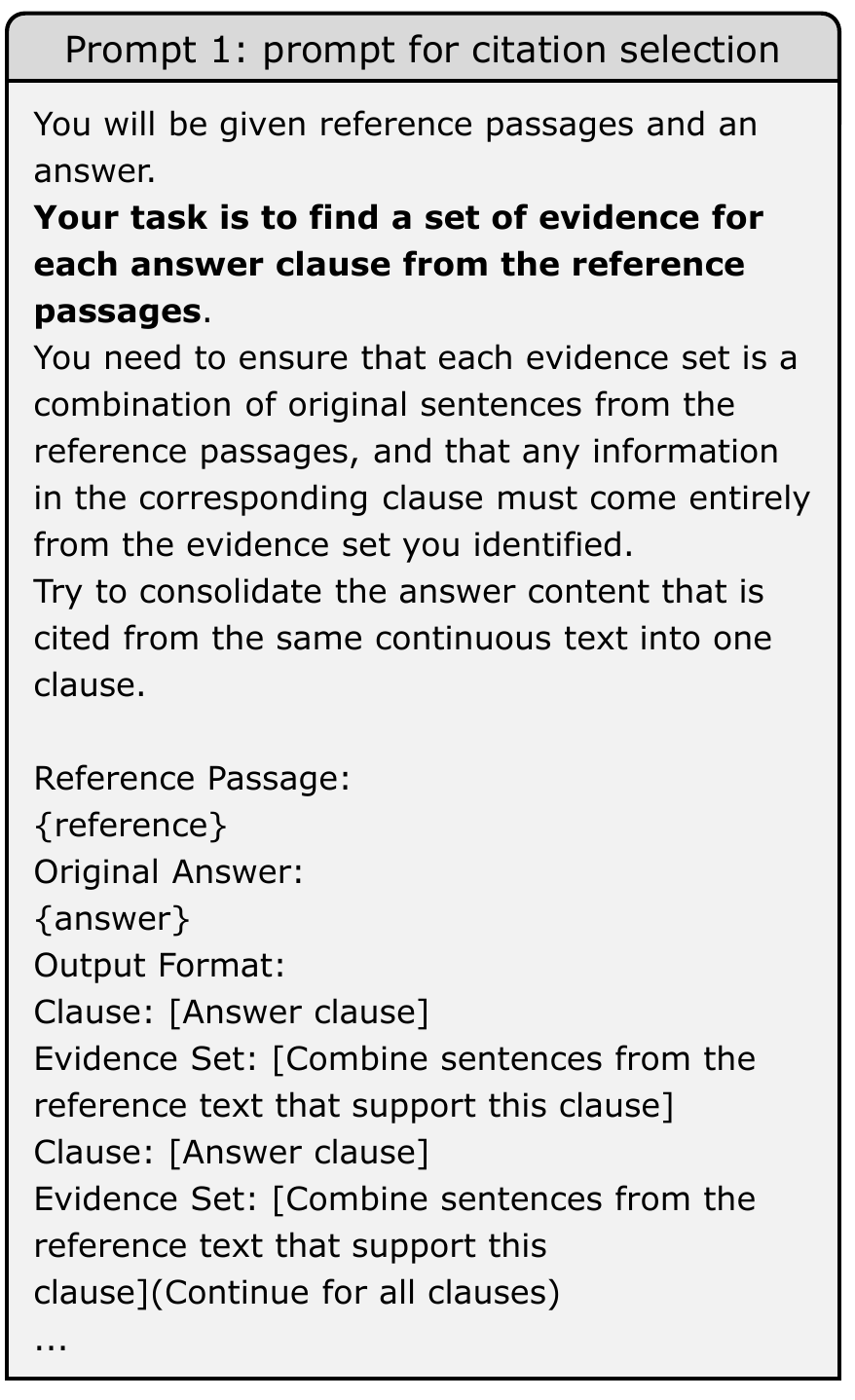}
    \label{fig:prompt1}
\end{figure}

\subsection{Citation Filtering}
After identifying the citation sentences corresponding to each answer clause, we need to filter out citation texts that diverge from the original reference passages or lack sufficient information to substantiate the answer clauses, the algorithm we use is as Algorithm~\ref{alg:algorithm1}.

\begin{algorithm}
\caption{Citation Filtering}\label{alg:algorithm1}
\begin{algorithmic}[1]
\State \textbf{Data}: Question $\mathcal{q}$, Reference passages $\mathcal{D}$, Answer $\mathcal{a} = \{(r_i, c_i) ~| ~r_i \in R, c_i \in C\}$, NLI model $M$
\For{each $Tuple (\mathcal{q}, \mathcal{D}, \mathcal{a})$}
    \State $flag \gets 1$
    \For{each citation $r_i$ and its corresponding clause $c_i$}
        \For {$sentence \in~r_i$}
            \If{sentence not in $\mathcal{D}$}
                \State $flag \gets 0$
            \EndIf
        \EndFor
        \State $m \gets p(M(r_i, c_i) = 1)$
        \If{$m < 0.8$}
            \State $flag \gets 0$
        \EndIf
    \EndFor
    \If{$flag = 0$}
        \State remove $Tuple (\mathcal{q}, \mathcal{D}, \mathcal{a})$
    \EndIf
\EndFor

\end{algorithmic}
\end{algorithm}

\section{Experiment Settings}
\label{sec:detail_experiment}

\subsection{Details of Datasets}
\begin{figure}[ht]
    \centering
    \includegraphics[width=0.48\textwidth]{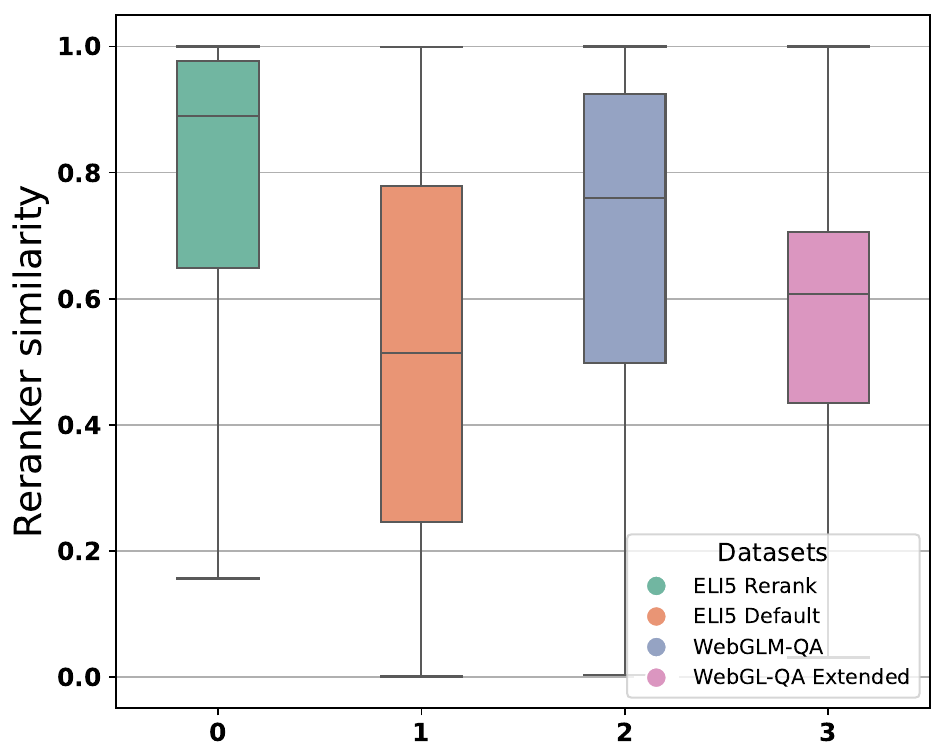}
    \caption{Reranker score distribution between query and reference passages in the training dataset.}
    \label{fig:train_similarity}
\end{figure}

\begin{figure}[ht]
    \centering
    \includegraphics[width=0.48\textwidth]{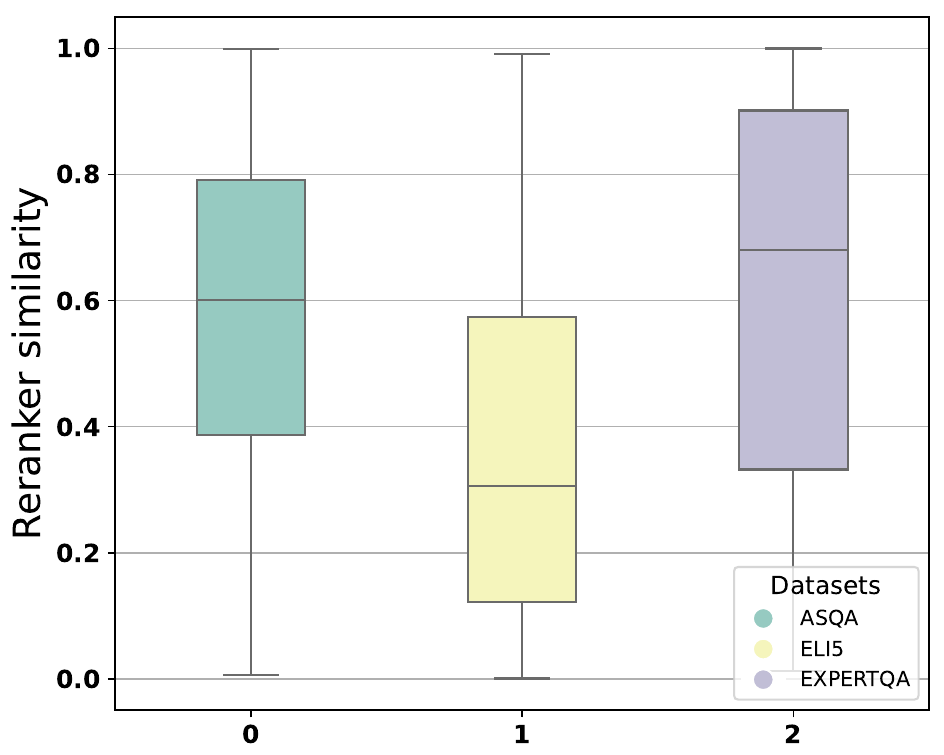}
    \caption{Reranker score distribution between query and reference passages in the test datasets.}
    \label{fig:test_similarity}
\end{figure}

The statistic of the test datasets is shown in Table~\ref{table:testdataset}. In Figure~\ref{fig:train_similarity} and Figure~\ref{fig:test_similarity}, we employ bge-v2-m3~\cite{li2023making,chen2024bge} to compute the relevance between queries and reference passages, presenting the statistical information of both training and test dataset in the form of box plots. Below, we provide a detailed description of the test datasets.

\paragraph{ASQA} ASQA \cite{stelmakh2022asqa} is the first long-form question answering dataset that focuses on ambiguous factoid questions (see Table~\ref{table:testdatacase}). It contains 4,353 samples for the train split and 948 samples for the dev split. For ALCE~\cite{gao2023enabling}, ASQA questions were paired with passages from Wikipedia passages (2018-12-20 snapshot) which purportedly contained the answers.

\paragraph{EXPERTQA} EXPERTQA~\cite{malaviya2024expertqa} dataset is a benchmark designed for evaluating the factuality and attribution capabilities of large language models across diverse domains. It contains 2177 long-form questions curated by experts from 32 fields, along with LLM-generated answers that have been revised by these experts to ensure factuality and proper sourcing. The dataset aims to provide a high-quality resource for developing and assessing AI systems that can deliver accurate and well-referenced information tailored to the needs of domain specialists.

\paragraph{ELI5} Descriptions can be found in the section~\ref{statistic_train_data}

\subsection{Details of Methods}
\paragraph{Prompting} In prompting setting, given a question and reference passages, we simply prompt the model to generate the attributed answer in a prescribed format. The prompt we use is as follows:

\begin{figure}[H]
    \centering
    \includegraphics[width=0.48\textwidth]{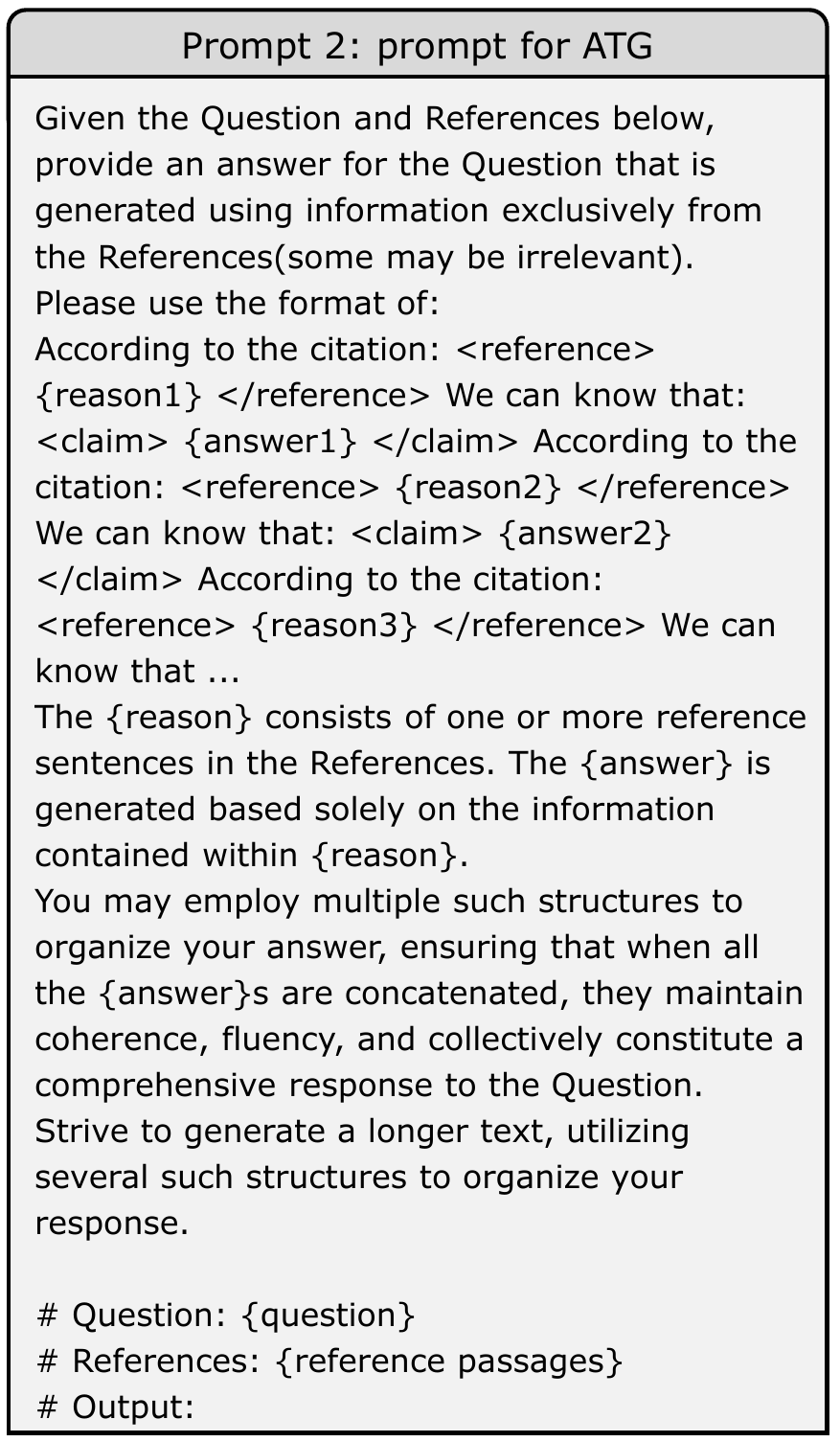}
    \label{fig:prompt2}
\end{figure}

\paragraph{\method$_\texttt{Unified}$} 
\label{section:unified_detail}
In \method$_\texttt{Unified}$ setting, we directly train the model using the constructed training dataset in Appendix~\ref{sec:detail_dataset} and employ the same instruction as in the prompting setting for one-step generation of attributed text.

\paragraph{\method~w/IG}
In \method~w/IG setting, we train two separate models for generating the reference and claim parts of attributed answers. Specifically, we train the model using the constructed training dataset in Appendix~\ref{sec:detail_dataset} as ReferModel for producing reference parts. For ClaimModel, which generates claim parts, we construct a new form of training data based on the original dataset. The detailed structure of this tailored training data is shown in Table~\ref{table:traindatacaseforclaim}. In Table~\ref{table:interleavingcase}, we provide an illustration of an interleaving generation instance.

\subsection{Details of Evaluation Metrics}
\label{sec:detail_metric}
\paragraph{EM Rec.} This metric is employed to assess the accuracy of the ASQA dataset. It utilizes a list of short answers (typically in the form of words or phrases) provided within the ASQA dataset to perform string matching within the generated long answers generated. The proportion of short answers that can be matched relative to the total length of the short answer list is calculated to determine the metric.

\paragraph{Claim Rec.} This metric is designed to evaluate the accuracy of the ELI5 and EXPERTQA datasets. Since these datasets do not provide short answer lists but only a long-form answer, ALCE employs InstructGPT~\cite{ouyang2022training} to sample three sub-claims from the standard answer. Then use the NLI model to assess the entailment relationship between each sub-claim and the LLM-generated response. The final score for this metric is determined by the proportion of sub-claims that are entailed in the LLM's answer.

\paragraph{MAUVE} We use MAUVE (Measuring the gap between neural text and human text using divergence frontiers) to measure the fluency of the generated text. MAUVE is a statistical metric that quantifies the similarity between neural-generated text and human-written text by computing the divergence frontiers between them. A higher MAUVE score indicates that the generated text is more coherent and natural, closely resembling human language style. 

\paragraph{CAS} This metric, consistent with the Citation Recall metric in ALCE, is designed to evaluate the proportion of answer clauses that are sufficiently supported by citations. Specifically, for each answer clause associated with a citation, a Natural Language Inference (NLI) model is employed to determine whether an entailment relationship exists between the clause and its corresponding citation. If such a relationship is identified, it indicates that the citation adequately substantiates the answer clause. The final score for this metric is derived from the ratio of answer clauses that exhibit an entailment relationship with their citations.

\paragraph{CRS} This metric aligns with the Citation Precision metric in ALCE, designed to assess whether the citations contain redundancy. In the ALCE, citations are identified by paragraph numbers, and redundancy is evaluated by sequentially removing paragraphs. However, in our approach, to accommodate sentence-level citations, we first segment the citation into sentences using NLTK. We then iteratively remove sentences and verify whether the remaining text still maintains an entailment relationship with the answer clause. Finally, the proportion of non-redundant citations is used as the CRS score.

\subsection{Training Details}
We trained the model using llama-factory~\cite{zheng2024llamafactory} on four A800 80G GPUs. For the Unified approach, we applied full fine-tuning with a learning rate of 3e-5, 3 epochs, and a total batch size of 128. For the \method~w/IG approach, we used LoRA to train two models: ReferModel with a learning rate of 5e-5, 5 epochs, and a total batch size of 128, and ClaimModel with a learning rate of 3e-5, 2 epochs, and a total batch size of 128. During model inference, we can alternate between two LoRA parameters to reduce memory requirements.

\section{Additional Results}
\subsection{Default Datasets}
For our test datasets, in addition to providing gold passages or reranked passages as reference passages, we also tested the results of directly using the top-5 retrieved passages, as shown in Table~\ref{table:defaultResults}.

\begin{table}[H]
\centering
\resizebox{0.48\textwidth}{!}{%
    \begin{tabular}{cccccc}
      \toprule
      \textbf{Dataset} & \textbf{Type} & \textbf{Fluency} & \textbf{Correct.} & \textbf{CAS} & \textbf{CRS} \\
      \midrule
      \multirow{2}{*}{ASQA} & Oracle & 88.1 & 53.5 & 92.1 & 71.6 \\
                             & Default & 88.1 & 40.0 & 91.0 & 84.4 \\
      \midrule
      \multirow{2}{*}{ELI5} & Oracle & 71.6 & 17.8 & 89.9 & 67.5 \\
                             & Default & 75.7 & 6.5 & 89.4 & 66.3 \\
      \midrule
      \multirow{2}{*}{EXPERTQA} & Rerank & 63.5 & 14.0 & 90.1 & 68.6 \\
                             & Default & 62.9 & 12.2 & 87.2 & 65.7 \\
      \bottomrule
    \end{tabular}
}
\caption{Results of the default test datasets.}
\label{table:defaultResults}
\end{table}

The results reveal that although the accuracy of the LLM's responses decreases due to varying quality of reference passages provided to the LLMs, it still maintains high CAS and CRS scores. This indicates that our method maintains strong attribution performance even with passages retrieved directly.

\subsection{Additional Baselines}
Due to the differences in datasets and evaluation metrics, we include this baseline in the Appendix. 

Learning to Plan and Generate Text with Citations~\cite{fierro-etal-2024-learning} conceptualizes plans as a sequence of questions that serve as blueprints for content generation and organization. Two variants of blueprint models are introduced: an abstractive model, where questions are generated from scratch, and an extractive model, where questions are copied from input passages.

For the comparative analysis of this method, we utilized the ASQA and ELI5 datasets, with top-5 reference passages retrieved directly via BM25 and GTR. The experimental results are shown in Table~\ref{table:lpgt}.

\begin{table}[H]
\centering
\resizebox{0.48\textwidth}{!}{%
    \begin{tabular}{ccccccc}
      \toprule
       & \multicolumn{2}{c}{\textbf{ASQA}} & \multicolumn{2}{c}{\textbf{ELI5}} & \multicolumn{2}{c}{\textbf{Average}} \\
       & \textbf{C} & \textbf{A} & \textbf{C} & \textbf{A} & \textbf{C} & \textbf{A} \\
      \midrule
      \makecell{LongT5 3B (10-psg)\\+$Blueprint_A$~\textup{+Attribution}} & 33.8 & 77.8 & 5.2 & 60.9 & 19.5 & 69.4 \\
      \midrule
      ~\method w/IG (5-psg) & 40.0 & 87.6 & 6.5 & \textbf{76.1} & 23.3 & \textbf{81.9} \\
      ~\method w/IG (10-psg) & \textbf{40.8} & \textbf{88.5} & \textbf{8.0} & 72.9 & \textbf{24.4} & 80.7 \\
      \bottomrule
    \end{tabular}
}
\caption{Comparison with the baseline: Learning to Plan and Generate Text with Citations. \textbf{C} denotes accuracy, while \textbf{A} represents the attribution score, which is the F1 score of CAS and CRS.}
\label{table:lpgt}
\end{table}

The experimental results indicate that our approach outperforms this method in terms of both answer accuracy and attribution scores.

\subsection{Faithfulness Results}
\begin{figure}[ht]
    \centering
    \includegraphics[width=0.48\textwidth]{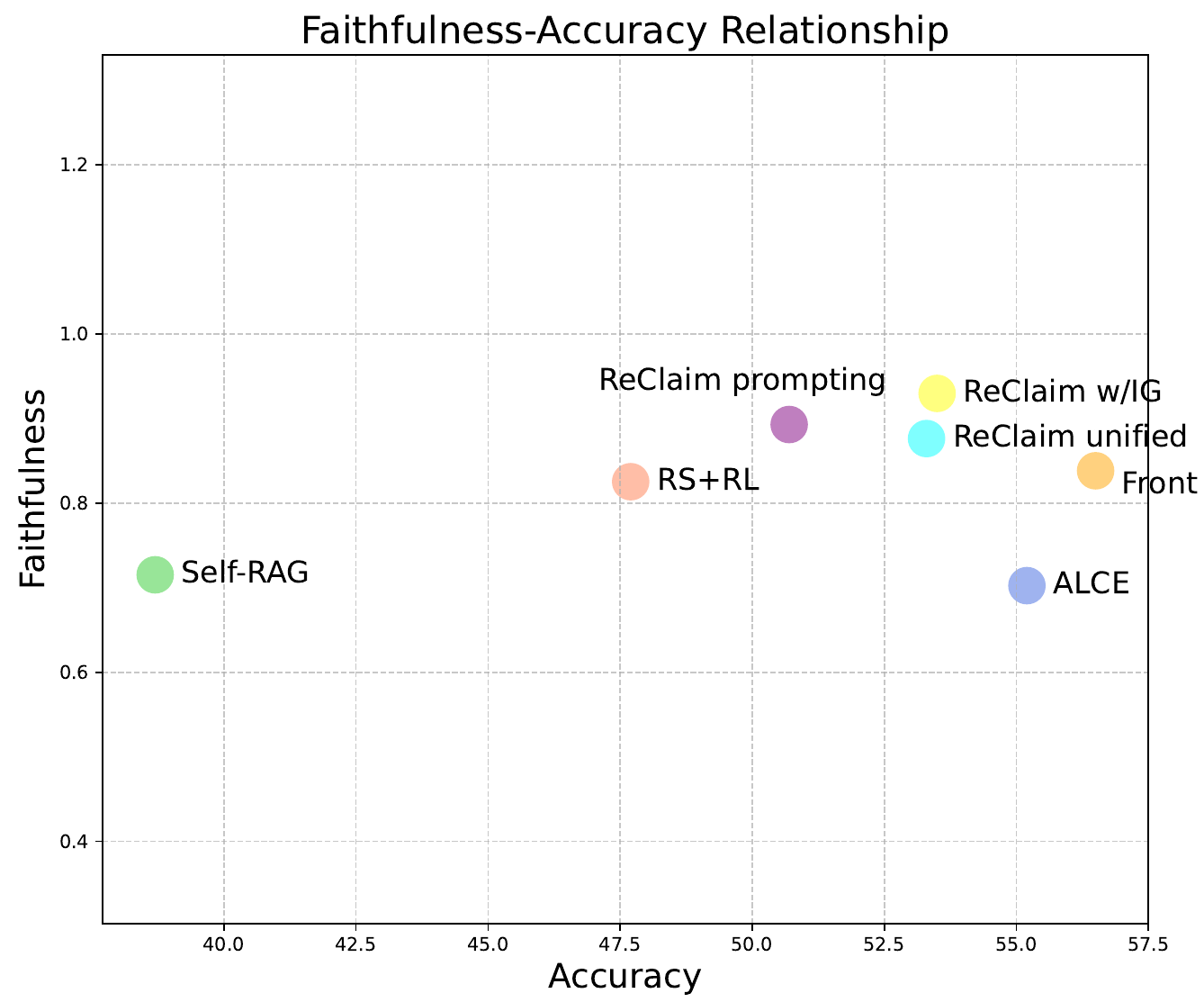}
    \caption{The faithfulness analysis results of the ASQA dataset.}
    \label{fig:faithfulness_asqa}
\end{figure}

\begin{figure}[ht]
    \centering
    \includegraphics[width=0.48\textwidth]{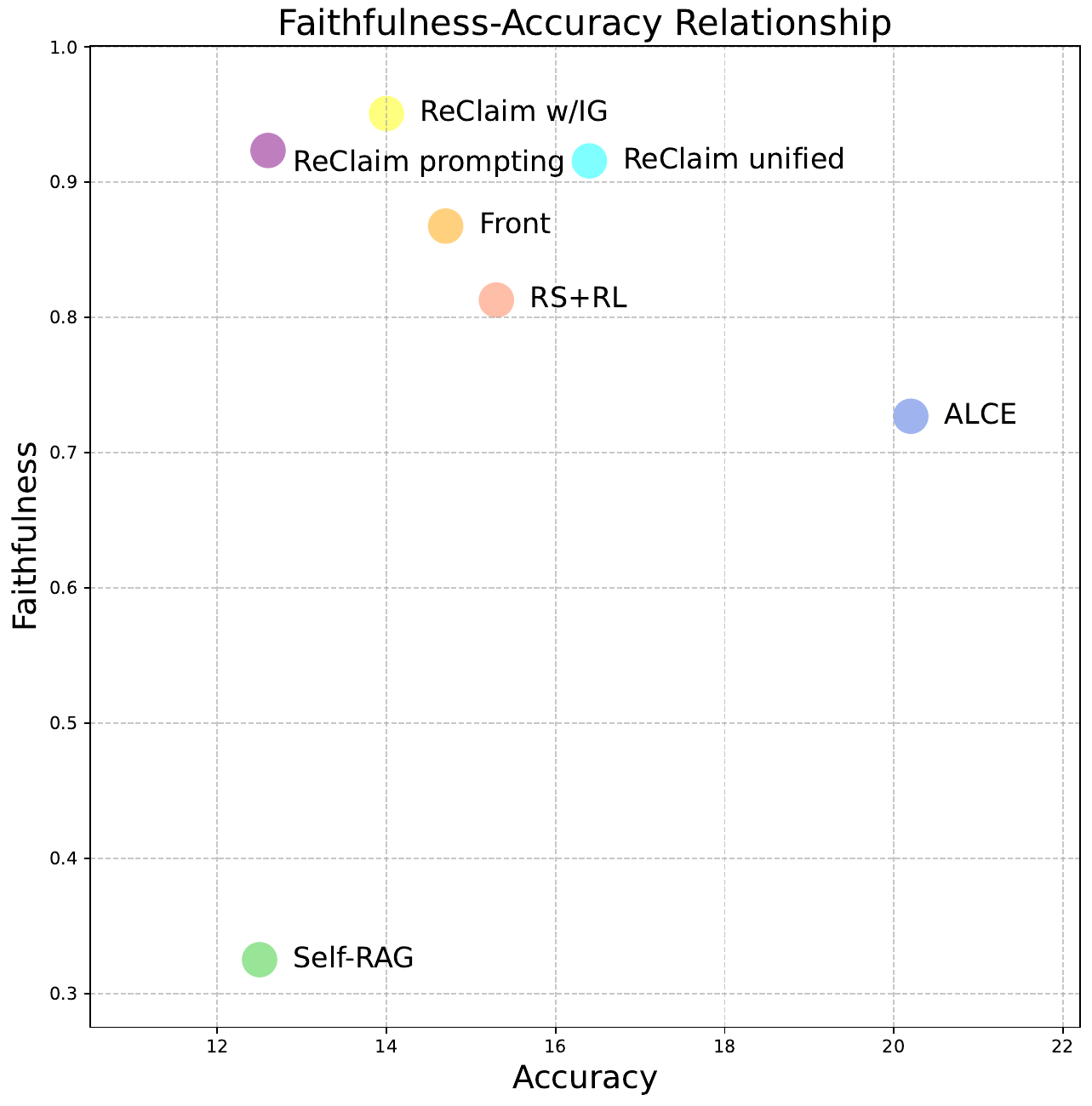}
    \caption{The faithfulness analysis results of the EXPERTQA dataset.}
    \label{fig:faithfulness_expertqa}
\end{figure}

The experimental results of the faithfulness metric for the ASQA and EXPERTQA datasets are shown in Figure~\ref{fig:faithfulness_asqa} and Figure~\ref{fig:faithfulness_expertqa}.

\begin{figure}[ht]
    \centering
    \includegraphics[width=0.48\textwidth]{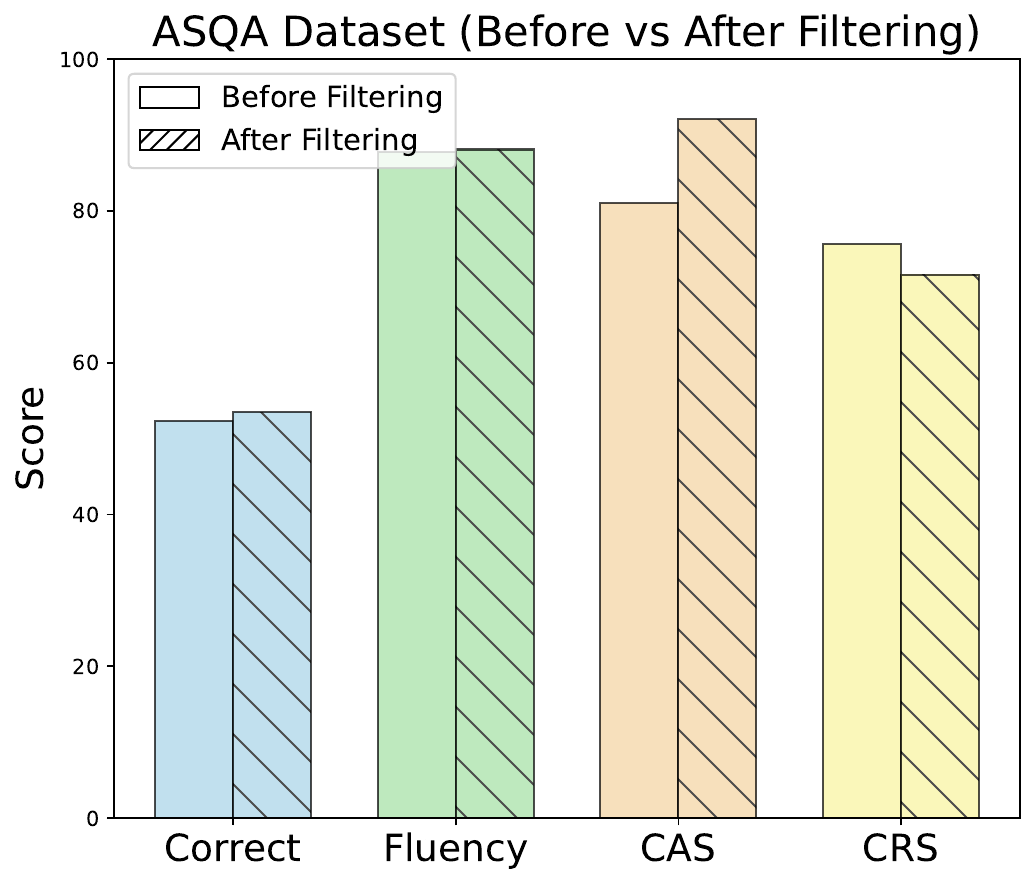}
    \caption{The comparative experimental results of the ASQA dataset.}
    \label{fig:data_filter_asqa}
\end{figure}

\begin{figure}[ht]
    \centering
    \includegraphics[width=0.48\textwidth]{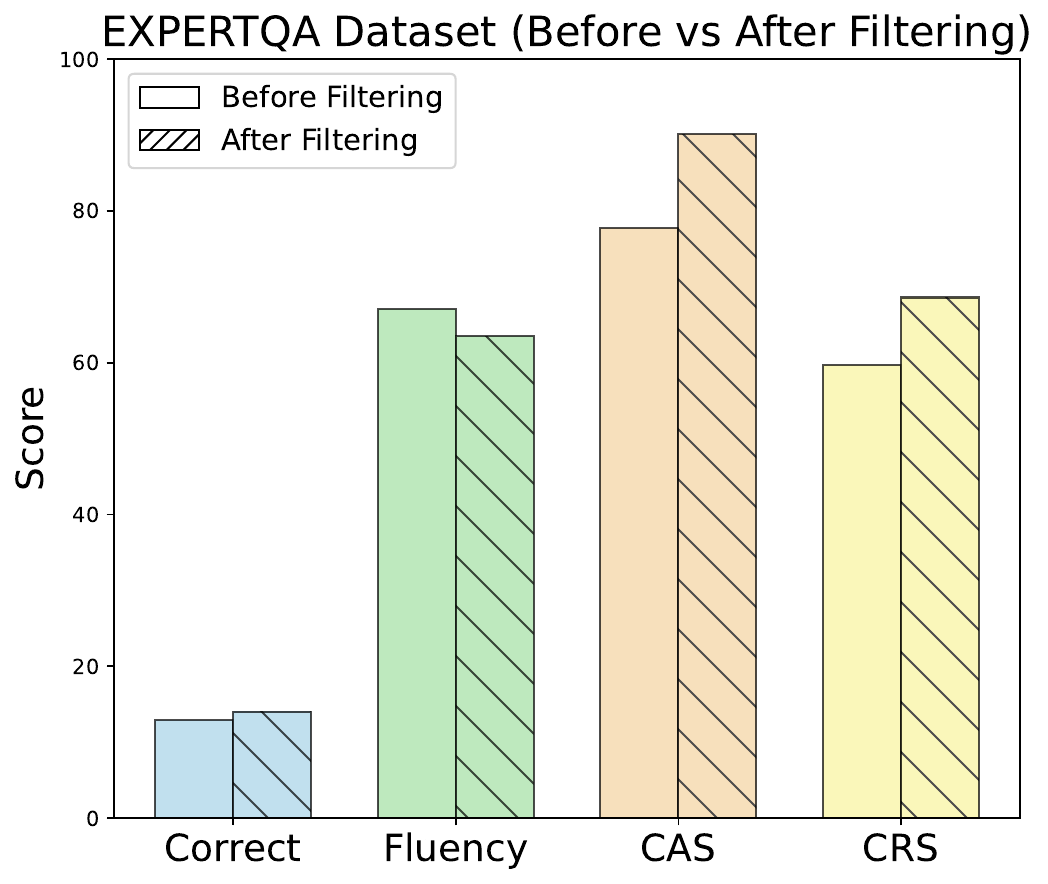}
    \caption{The comparative experimental results of the EXPERTQA dataset.}
    \label{fig:data_filter_expertqa}
\end{figure}


\begin{table*}[t]
\centering
\resizebox{0.98\textwidth}{!}{%
    \begin{tabular}{p{0.12\linewidth} p{0.86\linewidth}}
      \hline
        \textbf{Question} & Why does a dryer sheet help with my staticy shirt? \\ \hline
        \textbf{References} & [1] Dryer sheets are specifically designed to help reduce static cling in clothes by neutralizing the electric charges that build up during the drying process. Just toss a sheet in with your clothes, and they should come out nice and static-free when they’re done drying.\newline [2] So, if you if you want your clothes to be ready to wear, but they’re always covered in pet hair or are super clingy, then you definitely need to prevent static cling! Don’t worry, we’ve all been there. Bounce Dryer Sheets will help your clothes lead a no-strings-attached lifestyle in no time. Just toss a sheet into the dryer with your clothes, and leave that static cling behind!\newline [3] However, this coating can stick to your dryer and can be bad for the environment as well. Those looking to know how to reduce static on clothes without dryer sheets can find some solid, eco-friendly dryer sheet alternatives in many different places.\newline [4] The most common way people know how to prevent dryer static on clothes is with dryer sheets. Dryer sheets are sheets that are coated in a fabric softener full of positively charged electrons. These bond to the negatively charged ones and keep static from happening.\newline [5] Dryer static on clothes is one of the most annoying parts of doing laundry. It might seem like an unavoidable problem, but we have good news! There are some simple steps you can take to keep your laundry from becoming overly clingy or giving you static shocks. \\ \hline
        \textbf{Answer} & According to the citation: <reference> The most common way people know how to prevent dryer static on clothes is with dryer sheets. Dryer sheets are sheets that are coated in a fabric softener full of positively charged electrons. These bond to the negatively charged ones and keep static from happening. </reference> We can know that: <claim> Dryer sheets are coated in a fabric softener full of positively charged electrons, which bond to the negatively charged ones in your clothes and keep static from happening. </claim> \\ \hline
    \end{tabular}
}
\caption{Example of the training dataset.}
\label{table:traindatacase}
\end{table*}

\begin{table*}[t]
\centering
\resizebox{0.98\textwidth}{!}{%
    \begin{tabular}{ccccccc}
      \toprule
      \multirow{2}{*}{\textbf{Dataset}} & \multirow{2}{*}{\textbf{Samples}} & \multirow{2}{*}{\textbf{Question Type}} & \multirow{2}{*}{\textbf{\#passages}} & \multicolumn{3}{c}{\textbf{Average Length}}
      \\
      \cmidrule(lr){5-7}
      & & & & \multicolumn{1}{c}{\textbf{Question}} & \multicolumn{1}{c}{\textbf{Passages}} & \multicolumn{1}{c}{\textbf{Answer}} \\
      \midrule
      ASQA & 948 & Factoid (ambiguous) & Wikipedia (21M) & 9.0 & 517.5 & 71.8 \\
      ELi5 & 1000 & Why/How/What & Sphere (899M) & 16.5 & 546.3 & 121.5 \\
      EXPERTQA & 1000 & ambiguous/unambiguous & Sphere (899M) & 19.4 & 600.0 & 152.2 \\
       
      \bottomrule
    \end{tabular}
}
\caption{Statistics of the test datasets.}
\label{table:testdataset}
\end{table*}

\begin{table*}[t]
\centering
\resizebox{0.98\textwidth}{!}{%
    \begin{tabular}{p{0.12\linewidth} p{0.86\linewidth}}
      \hline
        \textbf{Input} & According to the citation: <reference> The most common way people know how to prevent dryer static on clothes is with dryer sheets. Dryer sheets are sheets that are coated in a fabric softener full of positively charged electrons. These bond to the negatively charged ones and keep static from happening. </reference> \\ \hline
        \textbf{Output} & We can know that: <claim> Dryer sheets are coated in a fabric softener full of positively charged electrons, which bond to the negatively charged ones in your clothes and keep static from happening. </claim> \\ \hline
    \end{tabular}
}
\caption{Example of the training dataset for claim generation.}
\label{table:traindatacaseforclaim}
\end{table*}

\onecolumn
\clearpage
\begin{longtable}{p{0.98\linewidth}}
    \hline \textbf{Question}: When data is compressed or zipped, what is actually happening to the data? \newline \newline
    \textbf{Reference Passages}: \newline 
    [1] Title: Lossless vs. Lossy Compression: What's the Difference? \newline 
    Text: data in several different ways, balancing fidelity and efficiency for functional and presentable data on the egress end. \textcolor{DarkGreen}{A common implementation of lossless file-compression includes the use of Huffman coding, whose redundancy-limiting algorithm recognizes patterns in groups in order to conserve time, space and other resources.} The model is able to compress and decompress digital media such that the output perfectly matches the input. The Zip file archival tool is a well-known format that supports lossless compression. Zip files compress digital media into much smaller data (often given the \u2018.zip\u2019 file extension) that can be uncompressed into their original form, \newline 
    [2] Title: What is MP3? (Designing Web Audio) \newline 
    Text: space, try encoding in mono. To make sure you trap all possible spatial data, encode in stereo mode. Most users find that joint stereo is adequate for most purposes. As mentioned earlier, compressing a WAV file with zip doesn't shave much off the file size, which is why psychoacoustics are employed. However, the MP3 encoding process actually does employ the classic Huffman encoding algorithm. \textcolor{DarkBlue}{After all psychoacoustic methods have been applied, the Huffman encoding pass seeks out and compresses any remaining redundancies in the bit pattern.} It's as though zip-type encoding were being run internally on the psychoacoustically encoded data \newline 
    [3] Title: US Treasury: addendum | Zip (File Format) | Comma Separated Values \newline 
    Text: most large files on the Internet are compressed. File compression reduces the size of a file and the time it takes to download. \textcolor{DarkYellow}{Compression software uses complex mathematical equations to scan a file for repeating patterns in the data. It replaces the data with smaller codes that take up less room.} This can dramatically decrease the size of report files. Viewing a \u201cZipped\u201d file To view compressed files, you need a compatible decompression program that can read these codes and convert the data back to its original form. To decompress a .zip file you need a utility like WinZIP\u2122 (an \newline 
    [4] Title: An introduction to data compression \newline 
    Text: data and try to find redundant parts of that data and remove those redundancies. That wasn't too simple, but the definition needs to be somewhat complex in order to differentiate compression from some other technologies. 2. How is compression accomplished? Can you give us a picture of what actually happens to the data? \textcolor{Brown}{Basically, they're trying to find common pieces of data blocks that they can get rid of, shrink, remove or substitute with smaller patterns. The more of those things they can find, the more it can compress.} A perfect example to illustrate how that works is if you \newline 
    [5] Title: Introduction to Data Compression - Data Compression | Coursera \newline
    Text: we get into the topic. so, just specific applications that, maybe are familiar for file compression. All file systems and, and disks have built-in, compression technologies. Such as, as zip, and b-zip and many others of similar type. Technologies. And the multimedia, everybody's familiar with, the JPEG and MP3 and MPEG, and all those sorts of things for images, sound and video. Those are all about data compression. and for communication, now, that's, what has, what enabled, fax, and also enables new technologies, like Skype, the ability to, reduce the, amount of data that you actually need to send. for \textcolor{red}{\textbf{(Irrelevant passage)}} \\
    \textbf{Gold Answer}: For most algorithms, the compressor will search for a pattern that appears multiple times within the data, and then replace that pattern with a shorter symbol, and create an entry in a "dictionary" recording what each symbol means. Compressing algorithms can get a little more complex than that, but that's the general idea. \newline \newline
    \textbf{Model Answer}: According to the citation: <reference> \textcolor{DarkYellow}{Compression software uses complex mathematical equations to scan a file for repeating patterns in the data. It replaces the data with smaller codes that take up less room.} </reference> We can know that: <claim> Compression software uses complex mathematical equations to scan a file for repeating patterns in the data and replaces the data with smaller codes that take up less room. </claim> According to the citation: <reference> \textcolor{DarkGreen}{A common implementation of lossless file-compression includes the use of Huffman coding, whose redundancy-limiting algorithm recognizes patterns in groups in order to conserve time, space and other resources.} </reference> We can know that: <claim> A common implementation of lossless file-compression includes the use of Huffman coding, whose redundancy-limiting algorithm recognizes patterns in groups in order to conserve time, space and other resources. </claim> According to the citation: <reference> \textcolor{DarkBlue}{After all psychoacoustic methods have been applied, the Huffman encoding pass seeks out and compresses any remaining redundancies in the bit pattern.} </reference> We can know that: <claim> After all psychoacoustic methods have been applied, the Huffman encoding pass seeks out and compresses any remaining redundancies in the bit pattern. </claim> According to the citation: <reference> \textcolor{Brown}{Basically, they're trying to find common pieces of data blocks that they can get rid of, shrink, remove or substitute with smaller patterns. The more of those things they can find, the more it can compress.} </reference> We can know that: <claim> The more of these common pieces of data blocks that can be found, the more it can compress. </claim>
    \\ \hline
    \caption{Example of the test data.} \label{table:testdatacase} \\
\end{longtable}
\twocolumn

\begin{table*}[ht]
\centering
\resizebox{0.98\textwidth}{!}{%
    \begin{tabular}{p{0.06\linewidth} p{0.12\linewidth} p{0.5\linewidth} p{0.3\linewidth}}
      \hline
        \textbf{Step} & \textbf{Model} & \textbf{Input} & \textbf{Output} \\ \hline
        1 & ReferModel & Instruction: Given the Question and References below, provide an answer for the Question that is generated using information exclusively from the References(some may be irrelevant). ... \newline \# Question: Who has the highest goals in world football? \newline \# References: "Football Player of the Century", by IFFHS International Federation of Football History and Statistics, 1999, "South America Football Player of the Century", by IFFHS International Federation of Football History and Statistics. Pelé's 1281 goals are recognized by FIFA as the highest total achieved by a professional footballer, although the Soccer Statistic Foundation (rssf) recognizes only 767 goals in official mode, occupying the third place after Josef Bican (805) and Romario (772). ... \newline Output: & According to the citation: <reference> Pelé's 1281 goals are recognized by FIFA as the highest total achieved by a professional footballer, although the Soccer Statistic Foundation (rssf) recognizes only 767 goals in official mode, occupying the third place after Josef Bican (805) and Romario (772). </reference> \\ \hline
        2 & ClaimModel & According to the citation: <reference> Pelé's 1281 goals are recognized by FIFA as the highest total achieved by a professional footballer, although the Soccer Statistic Foundation (rssf) recognizes only 767 goals in official mode, occupying the third place after Josef Bican (805) and Romario (772). </reference> & We can know that: <claim> Pelé's 1281 goals are recognized by FIFA as the highest total achieved by a professional footballer, although the Soccer Statistic Foundation (rssf) recognizes only 767 goals in official mode, occupying the third place after Josef Bican (805) and Romario (772). </claim> \\ \hline
        3 & ReferModel & Instruction: Given the Question and References below, provide an answer for the Question that is generated using information exclusively from the References(some may be irrelevant). ... \newline \# Question: Who has the highest goals in world football? \newline \# References: "Football Player of the Century", ... \newline Output: According to the citation: <reference> Pelé's 1281 goals are recognized by FIFA ... </reference> We can know that: <claim> Pelé's 1281 goals are recognized by FIFA ... </claim> & According to the citation: <reference> Rec.Sport.Soccer Statistics Foundation (RSSSF) estimates that he scored at least 805 goals in all competitive matches, which would make him the most prolific scorer of all time. </reference> \\ \hline
        4 & ClaimModel & According to the citation: <reference> Pelé's 1281 goals are recognized by FIFA ... </reference> We can know that: <claim> Pelé's 1281 goals are recognized by FIFA ... </claim> According to the citation: <reference> Rec.Sport.Soccer Statistics Foundation (RSSSF) estimates that he scored at least 805 goals in all competitive matches, which would make him the most prolific scorer of all time. </reference> & We can know that: <claim> Rec.Sport.Soccer Statistics Foundation (RSSSF) estimates that he scored at least 805 goals in all competitive matches, which would make him the most prolific scorer of all time. </claim> \\ \hline
        ... & ... & ... & ... \\ \hline
        
    \end{tabular}
}
\caption{Example of the interleaving generation.}
\label{table:interleavingcase}
\end{table*}

\end{document}